%% file: method_final.tex
\documentclass{article}

\PassOptionsToPackage{numbers, compress}{natbib}

\usepackage[final]{neurips_2022}
\usepackage{natbib}
\setcitestyle{square}
\setcitestyle{nonatbib}

\usepackage[utf8]{inputenc} 
\usepackage[T1]{fontenc}    
\usepackage{hyperref}       
\usepackage{url}            
\usepackage{booktabs}       
\usepackage{amsfonts}       
\usepackage{nicefrac}       
\usepackage{microtype}      
\usepackage{xcolor}         
\usepackage{graphicx} 
\usepackage{amssymb} 
\usepackage{amsmath} 
\usepackage{multirow} 
\usepackage{float} 
\usepackage{makecell} 
\usepackage{wrapfig} 
\makeatletter
\newcommand\figcaption{\def\@captype{figure}\caption}
\newcommand\tabcaption{\def\@captype{table}\caption}
\makeatother 

\input{xcl_math}

\title{Boosting Out-of-distribution Detection with\\ Typical Features}

\author{%
  Yao Zhu$^{1,2}$ \thanks{Yao Zhu is with the Zhejiang University, Hangzhou, China, 310013. (E-mail: ee$\_$zhuy$@$zju.edu.cn).} \quad Yuefeng Chen$^{2}$ \quad Chuanlong Xie$^{3}$ \quad Xiaodan Li$^{2}$ \quad Rong Zhang$^{2}$ \\ 
  \textbf{\quad Hui Xue$^{2}$ \quad Xiang Tian$^{1,5}$ \thanks{Corresponding authors: Xiang Tian, Bolun Zheng. (E-mail: tianx$@$zju.edu.cn, blzheng$@$hdu.edu.cn)} \quad Bolun Zheng$^{4,5}$ \quad Yaowu Chen$^{1,6}$} \\
  $^{1}$Zhejiang University, $^{2}$Alibaba Group, $^{3}$Beijing Normal University, $^{4}$Hangzhou Dianzi University\\ $^{5}$Zhejiang Provincial Key Laboratory for Network Multimedia Technologies \\
  $^{6}$ Zhejiang University Embedded System Engineering Research Center, Ministry of Education of China
}

\begin{document}

\maketitle

\begin{abstract}
Out-of-distribution (OOD) detection is a critical task for ensuring the reliability and safety of deep neural networks in real-world scenarios.
Different from most previous OOD detection methods that focus on designing OOD scores or introducing diverse outlier examples to retrain the model, we delve into the obstacle factors in OOD detection from the perspective of typicality and regard the feature's high-probability region of the deep model as the feature's typical set.
We propose to rectify the feature into its typical set and calculate the OOD score with the typical features to achieve reliable uncertainty estimation.
The feature rectification can be conducted as a {plug-and-play} module with various OOD scores.
We evaluate the superiority of our method on both the commonly used benchmark (CIFAR) and the more challenging high-resolution benchmark with large label space (ImageNet). Notably, our approach outperforms state-of-the-art methods by up to 5.11$\%$ in the average FPR95 on the ImageNet benchmark \footnote{The code will be available at \href{https://github.com/alibaba/easyrobust}{this https URL}}.  
\end{abstract}

\section{Introduction} 
Deep neural networks have been widely applied in various fields. Apart from the success of deep models,  predictive uncertainty is essential in safety-critical real-world scenarios such as autonomous driving \cite{geiger2012wedrive,huang2020surveydrive}, medical \cite{litjens2017surveymedical}, financial \cite{ozbayoglu2020deepfinancial}, etc. When encountering some examples that the deep model has not been exposed to during training, we hope the model raises an alert and hands them over to humans for safe handling. Such a challenge is usually referred to as out-of-distribution (OOD) detection and has gained significant research attention recently \cite{liu2020energy,hendrycks17baseline,ODIN,sun2021react}.

Most of the existing research \cite{liu2020energy,hendrycks17baseline,ODIN,thulasidasan2019mixup,papadopoulos2021outlier,Mahalanobis,huang2021importance} worked on designing suitable OOD scores for the pre-trained neural network, hoping to assign higher scores to the in-distribution (ID) examples and lower scores to the out-of-distribution (OOD) examples. However, these methods overlook the obstacle factors in OOD detection caused by the model's internal mechanisms. 
In this paper, we rethink the OOD detection from a perspective of feature typicality. We observed that the distribution of the deep features of the training dataset on different channels is approximately consistent with the Gaussian distribution (See examples in Appendix \ref{App:channels}). 
Accordingly, we divide these features into typical features (fall in the high-probability region) and extreme features (fall in the low-probability region).
Extreme features rarely appear in training and attract less attention from the classifier than the typical features. We hypothesize the classifier can model the typical features better than the extreme features, and the extreme features may lead to ambiguity and imprecise uncertainty estimation.
Given the potential negative impact, properly dealing with these extreme features is a key to improving the performance of OOD detection.

\begin{figure}[htbp]
\centering
\includegraphics[width=0.95\textwidth]{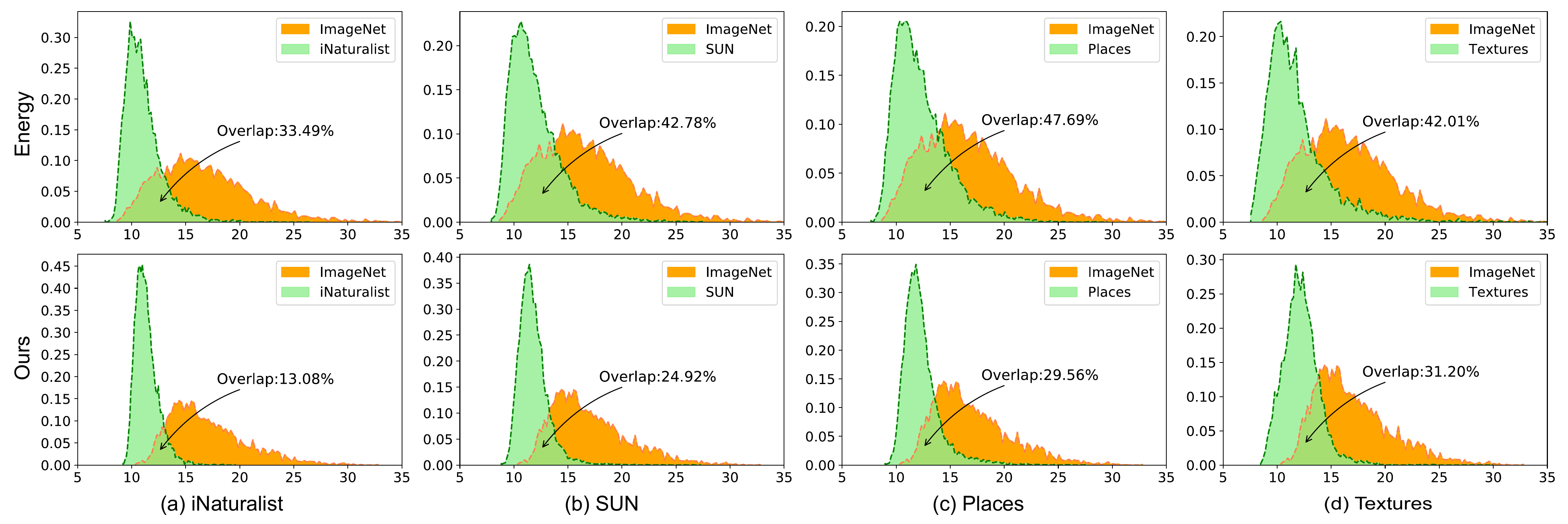}
\vspace{-0.4cm}
\caption{The distribution of the scores for ID (ImageNet) and OOD examples on ResNet-50. We use the energy score \cite{liu2020energy} as the OOD score. "Energy" means calculating the OOD score with the original features. "Ours" means calculating the OOD score with the typical features.}
\label{img:hist}
\end{figure}

In this paper, we propose to rectify the features into their typical set and then calculate the OOD score with these typical features. In this way, the model conservatively utilizes the typical features to make decisions and alleviates the damage caused by extreme features, which can be beneficial to the OOD scores derived from the pre-trained classifier \cite{liu2020energy,hendrycks17baseline,ODIN,huang2021importance}. Then the problem is how to estimate the feature's typical set on different channels since this requires a sufficient number of in-distribution examples and is time-consuming. Luckily, the commonly used operation Batch Normalization can shed light on a shortcut to selecting the feature's typical set and we name our approach \textbf{B}atch Normalization \textbf{A}ssisted \textbf{T}ypical \textbf{S}et Estimation (\textbf{BATS}). The Batch Normalization layer endeavors to normalize the features of the training dataset to Gaussian distributions, which can be used to estimate the typical set for ID features. Typical features are more common in training, while extreme features are rare, which leads to difficulties for the model to estimate extreme features well. We truncate the deep features with the guidance of the Batch Normalization, rectifying the extreme features to the boundary values of typical sets. We illustrate the distribution of the OOD scores for ID (ImageNet) and OOD (four different datasets) examples in Fig. \ref{img:hist}. 
Rectifying the features into the typical set with our \textbf{BATS} contributes to improving the separability between ID and OOD examples. 

Theoretically, we analyze the benefit of \textbf{BATS} and the bias-variance trade-off influenced by the strength of the hyperparameter. A proper strength of \textbf{BATS} contributes to improving the estimation accuracy of the reject region. Empirically, we perform extensive evaluations and establish superior performance on both the large-scale ImageNet benchmark and the commonly used CIFAR benchmarks. \textbf{BATS} outperforms the previous best method by a large margin, with up to a 5.11$\%$ reduction in the false positive rate (FPR95) and a 1.43$\%$ improvement in AUROC. Moreover, \textbf{BATS} can also slightly improve the test accuracy and robustness of the pre-trained models.
The main contributions of our paper are summarized as follows:
 \begin{itemize}
 \item[$\bullet$] We provide novel insights into OOD detection from the perspective of typicality and propose to rectify the features into the typical set. We design a concise and effective approach to select the feature's typical set named \textbf{B}atch Normalization \textbf{A}ssisted \textbf{T}ypical \textbf{S}et Estimation (\textbf{BATS}). 
 \item[$\bullet$] We provide theoretical analysis and empirical ablation on the benefit of \textbf{BATS} from the perspective of bias-variance trade-off to improve the understanding of our approach.
 \item[$\bullet$] Extensive experiments show that \textbf{BATS} establishes a state-of-the-art performance among post-hoc methods on a suite of OOD detection benchmarks. Moreover, \textbf{BATS} can boost the performance of various existing OOD scores with typical features.
 \end{itemize}

\section{Related work}

The literature related to OOD detection can be broadly grouped into the following themes: post-hoc detection methods \cite{liu2020energy,hendrycks17baseline,ODIN,sun2021react,Mahalanobis,huang2021importance,wang2021can}, confidence enhancement methods \cite{thulasidasan2019mixup,papadopoulos2021outlier,hein2019relu,bitterwolf2020certifiably,hendrycks*2020augmix,yun2019cutmix,hendrycks2018deep,chen2021atom}, and density-based methods \cite{kobyzev2020normalizing,zisselman2020deep,serra2019input,xiao2020likelihood,nalisnick2018deep,kirichenko2020normalizing}. 
Post-hoc detection methods focus on improving the OOD uncertainty estimation by utilizing the pre-trained classifiers rather than retraining a model, which is beneficial for adopting OOD detection in real-world scenarios and large-scale settings. 
MSP \cite{hendrycks17baseline} observes that the maximum softmax probability of ID examples can be higher than that of the OOD examples and provide a simple baseline for OOD detection. ODIN \cite{ODIN} introduces a {sufficiently large} temperature factor and input perturbation to separate the ID and OOD examples. \citet{liu2020energy} analyze the limitations of softmax function in OOD detection and propose to use energy score as an indicator. The examples with high energy are considered OOD examples, and vice versa.
ReAct \cite{sun2021react} hypothesizes that the OOD examples can trigger the abnormal activation of the model and propose to clamp the activation value larger than the threshold to improve the detection performance.
GradNorm \cite{huang2021importance} shows that the gradients of the categorical cross-entropy loss can be an effective test statistic for OOD detection.
{Different from these methods, our \textbf{BATS} proposes to calculate the OOD scores with the typical features, which benefits the estimation of the reject region and can improve the detection performance.} 

\section{Preliminaries}

\subsection{Out-of-distribution detection}\label{testOOD}
In this section, we provide a summary of the out-of-distribution detection from the perspective of hypothesis testing \cite{nalisnick2019detecting,ahmadianlikelihood,haroush2021statistical,zhang2021understanding,bergamin2022model}.
We consider a classification problem with $K$ classes and denote the labels as $\cY = \{1,2,\ldots, K\}$.
Let $\cX$ be the input space. Suppose that the in-distribution data $\cD_{in}=\{(x_i, y_i)\}_{i=1}^n$ is drawn from a joint distribution $P_{X,Y}$ defined over $\cX \times \cY.$
We denote the marginal distribution of $P_{X,Y}$ for the input variable $X$ by $P_0.$ 
Given a test input $\rvx\in\cX$, the problem of out-of-distribution  detection can be formulated as a single-sample hypothesis testing task:
\benr \label{Eq:test}
\cH_0: \rvx \sim P_0, \quad \text{vs.} \quad \cH_1: \rvx \nsim P_0. 
\label{eq1}
\eenr
Here the null hypothesis $\cH_0$ implies that the test input $\rvx$ is an in-distribution sample.
The goal of OOD detection here is to design criteria based on $\cD_{in}$ to determine whether $\cH_0$ should be rejected. OOD detection tasks need to determine a reject region $\cR$ such that for any test input $\rvx \in \cX$, the null hypothesis is rejected if $x \in \cR.$
Generally, the reject region $\cR$ is formulated by a test statistic and a threshold.
Let $f: \cX \mapsto \R^K$ be a  model pre-trained from $\cD_{in}$, which is used to predict the class label of an input sample.   
One can use the model $f$ or a part of $f$ (e.g., feature extractor) to construct a test statistic $T(\rvx; f)$, where $\rvx$ is the test input. Then the reject region can be written as
$
\cR = \{\rvx: T(\rvx;f) \leq \gamma\}
$,
where $\gamma$ is the threshold.

\subsection{OOD detection with energy score}

For a classifier $f$ and a data point $(\rvx, y)$, we use $f(\rvx)[k]$ to represent the $k^{th}$ output of the last layer. With reference to~\cite{liu2020energy,grathwohl2019your,energyrobust}, the marginal density $p({\boldsymbol{x}})$ of the classifier can be expressed as: $p(x) = \frac{\exp(-E(x))}{Z} = \frac{\sum_{k=1}^{K} \exp({f(\rvx)[k]})}{Z}$, 
where $Z$ is the normalizing factor and is independent to $\rvx$. $E(x)$ represents the energy of $x$ and is modeled by neural network as $E(\rvx)=-\log \sum_{k=1}^{K} \exp({f(\rvx)[k]})$. See Appendix \ref{App:energy} for details.
Considering that $Z$ is a constant and is independent to $\rvx$, \citet{liu2020energy} propose an energy score that uses the opposite of the energy $E(\rvx)$ as a test statistic to detect OOD examples. A higher energy score means a higher marginal density $p({\boldsymbol{x}})$. 

\section{Methods}

In this paper, we delve into the obstacle factor for the post-hoc OOD detection from the perspective of typicality, which aims to boost the performance of the existing OOD scores and is orthogonal to the methods of designing different OOD scores. Given that the energy score \cite{liu2020energy} is {provably} aligned with the density of inputs and performs well, we mainly use the energy score as the OOD score. (See Appendix \ref{App:combine} for other OOD scores).


\subsection{Motivation}

For a classifier trained on the ID data $f = f_{w, b} \circ g$ where $g$ is a feature extractor mapping input $\mathbf{x}$ to its deep feature $\mathbf{z}$.
Let $d$-dimensional vector $\mathbf{z} = [z_{1}, ...,z_{d}]^{\top} = g(\mathbf{x})$ denote the deep features of $\mathbf{x}$ extracted by $g$, and $z_{i}$ indicate the $i$-th element of $\mathbf{z}$. $f_{\mathbf{w},\mathbf{b}}(\mathbf{z}) = \mathbf{w} \cdot \mathbf{z} + \mathbf{b} $ is a fully connected layer mapping the deep feature $\mathbf{z}$ to output logits.
The energy can be expressed as:
\benr
E(\rvx) = -\log \sum_{k=1}^{K} \exp({f(\rvx)[k]})=-\log \sum_{k=1}^{K} \exp({( \mathbf{w} \cdot \mathbf{z} + \mathbf{b})[k]}).
\eenr

The test statistic can be expressed as $T(\rvx; f) := -E(\rvx)=\log \sum_{k=1}^{K} \exp({( \mathbf{w} \cdot \mathbf{z} + \mathbf{b})[k]})$, which depends on the extracted deep features and the mapping operation of the fully connected layer (FC). Assuming that the distribution of the deep features is consistent with the Gaussian distribution (see examples in Appendix \ref{App:channels}), there are high-probability regions and low-probability regions in deep features. We name the features that fall in high-probability regions as typical features, and the corresponding regions are called feature's typical sets. In contrast, we regard the features that fall in low-probability regions as extreme features. Extreme features are rarely exposed to the training process, which leads to difficulties for the classifier to model these features and unreliable estimations in the inference process. Reducing the influence of extreme features on test statistics can be a key to improving OOD detection performance.

\subsection{Batch Normalization Assisted Typical Set Estimation} \label{BATS}
Instead of designing new OOD scores to detect the abnormality, we provide a novel insight into OOD detection from a perspective of typicality. We propose to rectify the features into the feature's typical set and then use these typical features to calculate the OOD score.
Consider a {commonly} used layer structure in deep convolutional networks:
\benr\label{layer}
\rvz' \rarrow \text{BN}(\rvz'; \mu, \sigma) \rarrow \text{ReLU}  \rarrow \rvz,
\eenr
where $\rvz'$ is the feature vector extracted from the convolutional layer of $g.$
To identify the typical set of $\rvz'$ for each channel, we should apply its feature map to a sufficient number of ID examples and further calculate the empirical distribution of $\rvz'$ over the ID examples. 
If the number of features is large, the inference procedure is time-consuming.
{\it Here we propose a simple and effective post hoc approach that leverages the information stored in $f$ to infer the typical set without estimating the distribution of $\rvz'.$}
Suppose the pre-trained deep neural network uses batch normalization (BN). We denote the BN unit in $f$ as:
\benr
\text{BN}(\rvz'; \mu, \sigma) = \sigma \frac{\rvz' - \E(\rvz')}{\text{Std}(\rvz')} + \mu,
\eenr
where  $\mu$, $\sigma$ are two learnable parameters.   
After the pre-training, all the four parameters $\mu$, $\sigma$, $\E(\rvz')$, $\text{Std}(\rvz')$ are known and stored in the weights of $f.$\footnote{In general, $\E(\rvz')$ and $\text{Std}(\rvz')$ are estimated on a mini-batch of the training data. Finally, the pre-trained model outputs moving average estimators at each iteration.}
The Batch Normalization normalizes features of the training dataset to a distribution with a mean of $\mu$ and standard deviation of $\sigma$, which means that the features fall in the interval $[\mu-\lambda*\sigma, \mu+\lambda*\sigma]$ appear more frequently in training than the features in the complement of this interval. {The parameter $\lambda$ controls the range of the interval.} Thus we use the information in the Batch Normalization to identify the in-distribution feature's typical set and rectify the features into the typical set before calculating the OOD score. The uncertainty estimated with the typical features can be more reliable.

In practice, we propose a truncated activation scheme to bound the output features of the BN unit.
First, we introduce the truncated BN unit by:

\benr
\text{TrBN}(\rvz'; \mu, \sigma, \lambda) =  \begin{cases}
\mu + \lambda \sigma, & \text{if} \quad \rvz' - \mu \geq \lambda \sigma; \\
\text{BN}(\rvz'; \mu, \sigma),  & \text{if} \quad -\lambda \sigma < \rvz' - \mu < \lambda \sigma  ; \\
\mu - \lambda \sigma, & \text{if}\quad  \rvz' - \mu \leq  -\lambda \sigma,
\end{cases}
\eenr

where $\lambda$ is a tuning parameter.
We replace the BN unit in the layer structure (Eq.(\ref{layer})) with the TrBN unit and write the rectified final features as $\bar {\mathbf{z}}$ and the new classifier as $\bar f$. 
Then test statistic with the energy score can be expressed as:
\benr
T(\rvx; \bar f) = \log \sum_{k=1}^{K} \exp({(\mathbf{w}\cdot \bar{\mathbf{z}}+\mathbf{b})}[k])
\eenr
and take the reject region by $\cR=\{\rvx: T(\rvx; \bar f)\leq\gamma\}.$ 
We name our approach \textbf{B}atch Normalization \textbf{A}ssisted \textbf{T}ypical \textbf{S}et Estimation (BATS). In comparison to the standard BN, the outputs of TrBN are concentrated toward the feature's typical set of ID data. This makes an ID example less susceptible to being mistakenly detected as an OOD example and buffers the negative impact of the extreme features. Fig. \ref{img:hist} compares the distribution of the OOD scores from the original energy score and the energy score with typical features.

\subsection{Theoretical analysis} \label{sec43}

The truncation threshold $\lambda$ is a key hyperparameter. 
Our method reduces the variance of $\rvz'$ and also introduces a bias term since it changes the distribution of $\text{BN}(\rvz'; \mu, \sigma).$ 
The variance reduction means that our method is robust to the rare ID examples, while the introduced bias can lead to degradation of the model performance.
In this section, we assume $\rvz'$ follows a normal distribution and analyze the bias-variance trade-off in our method. 


\subsubsection{Understanding the benefits of BATS from the perspective of variance reduction}
The variance reduction happens at the BN step. The variance of $\text{BN}(\rvz'; \mu, \sigma)$ is $\sigma^2$ since the distribution of the ID features $\rvz'$ is rescaled to $N(\mu, \sigma^2).$
While the TrBN unit truncates the extreme values and the variance of $\text{TrBN}(\rvz'; \mu, \sigma, \lambda)$ becomes:
\benr\label{var}
\sigma^2 C(\lambda):= \sigma^2 \Big( \text{erf}(\frac{\lambda}{\sqrt 2}) - \frac{\sqrt 2}{\sqrt \pi}\lambda \exp(-\frac{\lambda^2}{2}) + \lambda^2(1-\text{erf}(\frac{\lambda}{\sqrt 2})) \Big),
\eenr
where $\text{erf}(x) = (2/\sqrt \pi) \int_0^x \exp(-t^2) dt$ is the Gauss error function.
The value of $C(\lambda)$ represents the degree of variance reduction.
In Eq.(\ref{var}), $C(0)=0$, $d C(\lambda) / d \lambda >0$, and $C(\lambda) \to 1$ as $\lambda \to +\infty.$ Therefore, $C(\lambda)$ is a monotonically increasing function and $0\leq C(\lambda)<1$ for $0\leq \lambda < +\infty.$ In summary, the smaller $\lambda$, the smaller the variance. See Appendix \ref{App:proof} for the proof.

OOD detection is a single-sample hypothesis testing problem (in Eq.(\ref{Eq:test})), and the in-distribution $P_0$ is unknown.
So the reject region is determined by the empirical distribution of the test statistic $T(\rvx; f)$ over the ID data.
The extreme features increase the uncertainty and lead to more unusual values of $T(\rvx; f).$
This implies that the reject region may be underestimated due to the heavy tail property of $T(\rvx; f).$
Our \textbf{BATS} aids this problem by reducing the variance of the deep features, which contributes to constraining the uncertainty of $f$ and $T(\rvx; f)$ and improving the estimation accuracy of the reject region. 

\subsubsection{The bias introduced by BATS}
\textbf{BATS} rectifies the features into the typical set, which reduces the variance of the deep features. However, this operation can also introduce a bias term, which can reflect the change in the distribution of the features. A large bias can damage the performance of the model.  
The distribution of the output feature $\rvz = \text{ReLU}(\text{BN}(\rvz'; \mu, \sigma, \lambda))$ is a one-side rectified normal distribution over $[0, +\infty).$ The expectation of $\rvz$ is:
\benr
\E(\rvz) = \mu + \sigma \Big(\frac{1}{\sqrt{2\pi}} \big( \exp(-\frac{\mu^2}{2\sigma^2})  \big)   - \frac{\mu}{2\sigma}(1+\text{erf}(-\frac{\mu}{\sqrt{2}\sigma})) \Big).
\eenr
For our method, the distribution of the output feature $\bar \rvz = \text{ReLU}(\text{TrBN}(\rvz; \mu, \sigma, \lambda))$ is a {two-sided} rectified normal distribution over $[0, \mu+\lambda\sigma]$ and the expectation of $\bar \rvz$ is:
\benr
\E(\bar \rvz) = \mu + \sigma \Big(\frac{1}{\sqrt{2\pi}} \big( \exp(-\frac{\mu^2}{2\sigma^2}) - \exp(-\frac{\lambda^2}{2})  \big)   - \frac{\mu}{2\sigma}(1+\text{erf}(-\frac{\mu}{\sqrt{2}\sigma})) + \frac{\lambda}{2} (1-\text{erf}(\frac{\lambda}{\sqrt 2}))\Big).
\eenr
Then the bias caused by the truncation is:
\benr
\E(\bar \rvz)-\E(\rvz) = \sigma\Big( - \exp(-\frac{\lambda^2}{2}) + \frac{\lambda}{2} (1-\text{erf}(\frac{\lambda}{\sqrt 2})) \Big) = \big(\lambda - \lambda \Phi(\lambda) - \phi(\lambda)\big)\sigma,
\eenr
where $\phi(\cdot)$ and $\Phi(\cdot)$ are the probability density function (pdf) and cumulative distribution function (cdf) of the standard normal distribution. One can find that the bias term $\E(\bar \rvz)-\E(\rvz)$ converges to zero as $\lambda \to \infty.$ In other words, if $\lambda$ is large enough, the bias can be very small. Thus, there exists a bias-variance trade-off. See Appendix \ref{App:proof} for the proof.

A proper selection of $\lambda$ can improve the detection performance by significantly reducing the uncertainty (variance reduction) and slightly changing the distribution of the features (small bias). 
If $\lambda$ is large, $T(\rvx; \bar f)$ uses more extreme features in both the ID and OOD data.
As $\lambda$ tends to infinity,  $T(\rvx; \bar f)$ converges to $T(\rvx; f).$
Then \textbf{BATS} is the same to the original energy detection.
If $\lambda$ is small, extreme features are removed from the test statistic $T(\rvx; \bar f)$ while introducing a non-negligible bias.
Because of the change in feature distribution, the detection method loses its power to identify OOD examples.

\begin{figure}[htbp]
\centering
\includegraphics[width=0.95\textwidth]{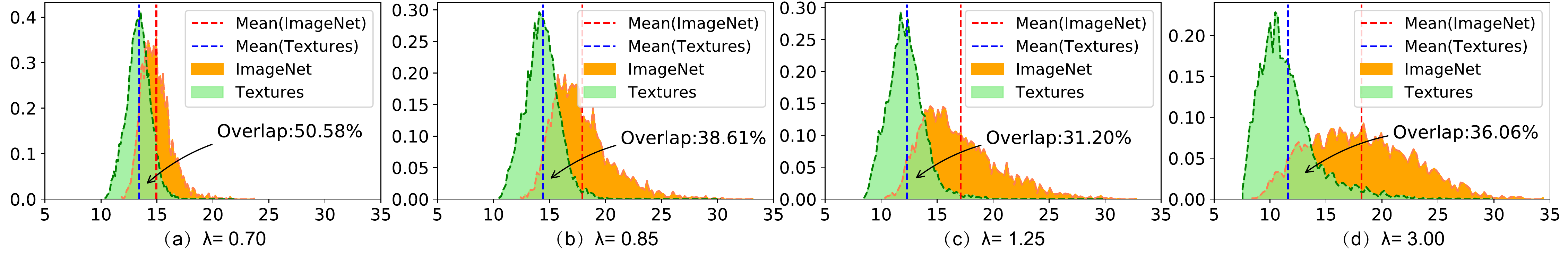}
\caption{Bias-variance trade-off in BATS. We illustrate the OOD score for ID (ImageNet) and OOD (Textures) examples. Smaller $\lambda$ contributes to variance reduction which benefits the estimation of the reject region. But smaller $\lambda$ causes a larger bias, which can drastically alter the distribution of features and damage the performance of the model in distinguishing ID and OOD examples.}
\vspace{-0.3cm}
\label{img:tradeoff}
\end{figure}

Fig. \ref{img:tradeoff} illustrates the distribution of OOD scores with different $\lambda$, which empirically verifies this trade-off.

\section{Experiments}
In this section, we first introduce our experiment implementation. Then, we evaluate our methods both on the large-scale OOD detection benchmark \cite{huang2021mos} and the CIFAR benchmarks \cite{hendrycks17baseline}. After that, the ablation studies compare the influence of applying rectification on different layers and show the influence of the hyperparameter. Moreover, our \textbf{BATS} can also slightly improve the test accuracy of the pre-trained models (in Appendix \ref{App:testacc}). We consider the out-of-distribution detection as a \textbf{single-sample} hypothesis testing task and only test one sample at a time.
\subsection{Implementation}
\textbf{Dataset.} 
For evaluating the large-scale OOD detection performance, we use ImageNet-1k \cite{huang2021mos} as the in-distribution dataset and consider four out-of-distribution datasets, including (subsets of) the fine-grained dataset iNaturalist \cite{van2018inaturalist}, the scene recognition datasets Places \cite{zhou2017places} and SUN \cite{xiao2010sun}, and the texture dataset Textures \cite{cimpoi2014Texture} with non-overlapping categories to ImageNet-1k. 

As for the evaluation on CIFAR Benchmarks, we use the CIFAR-10 and CIFAR-100 \cite{krizhevsky2009learningCIFAR} as the in-distribution datasets using the standard split with 50,000 training images and 10,000 test images. We consider four OOD datasets: SVHN \cite{netzer2011readingSVHN}, Tiny ImageNet \cite{chrabaszcz2017downsampledTiny}, LSUN \cite{yu2015lsun} and Textures \cite{cimpoi2014Texture}. 

\textbf{Baselines.} We consider different kinds of competitive OOD detection methods as baselines, including Maximum Softmax Probability (MSP) \cite{hendrycks17baseline}, ODIN \cite{ODIN}, Energy \cite{liu2020energy}, Mahalanobis \cite{Mahalanobis}, GradNorm \cite{huang2021importance} and ReAct \cite{sun2021react}. MSP is a simple baseline for OOD detection and ReAct is a state-of-the-art method that achieves strong detection performance. All methods use the pre-trained networks post-hoc.

\textbf{Metrics.} \textbf{FPR95:} the false positive rate of OOD (negative) examples when the true positive rate of in-distribution (positive) examples is as high as 95$\%$. Lower FPR95 indicates better OOD detection performance and {vice versa}. \textbf{AUROC:} the area under the receiver operating characteristic curve (ROC). Higher AUROC indicates better detection performance.
See Appendix \ref{App:details} for more details.

\begin{table}[htbp]
\caption{OOD detection performance comparison on different architectures: ResNet-50 (RN50) \cite{he2016resnet}, DenseNet-121 (DN121) \cite{huang2018densenet} and MobileNet-V2 (MNet) \cite{sandler2019mobilenetv2}. We use the pre-trained models in PyTorch \cite{pytorch} trained on ImageNet. All methods are post hoc and can be directly used for pre-trained models. The best results are in Bold. The up arrow indicates that the higher the value, the better the performance, and vice versa.} 
\scalebox{0.72}{
\begin{tabular}{cccccccccccc}
\hline
\multirow{3}{*}{Model} & \multirow{3}{*}{Method} & \multicolumn{2}{c}{iNaturalist} & \multicolumn{2}{c}{SUN} & \multicolumn{2}{c}{Places} & \multicolumn{2}{c}{Textures} & \multicolumn{2}{c}{Average} \\ 
 &  & FPR95 & AUROC & FPR95 & AUROC & FPR95 & AUROC & FPR95 & AUROC & FPR95 & AUROC \\
  &  & $\downarrow$ & $\uparrow$ & $\downarrow$ & $\uparrow$ & $\downarrow$ & $\uparrow$ & $\downarrow$ & $\uparrow$ & $\downarrow$ & $\uparrow$\\ \hline
\multirow{7}{*}{RN50} & MSP\cite{hendrycks17baseline} & 51.44 & 88.17 & 72.04 & 79.95 & 74.34 & 78.84 & 54.90 & 78.69 & 63.18 & 81.41 \\ 
 & ODIN\cite{ODIN} & 41.07 & 91.32 & 64.63 & 84.71 & 68.36 & 81.95 & 50.55 & 85.77 & 56.15 & 85.94 \\ 
 & Energy\cite{liu2020energy} & 46.65 & 91.32 & 61.96 & 84.88 & 67.97 & 82.21 & 56.06 & 84.88 & 58.16 & 85.82 \\ 
 & Mahalanobis\cite{Mahalanobis} & 97.00 & 52.65 & 98.50 & 42.41 & 98.40 & 41.79 & 55.80 & 85.01 & 87.43 & 55.47 \\  
 & GradNorm\cite{huang2021importance} & 23.73 & 93.97 & 42.81 & 87.26 & 55.62 & 81.85 & \textbf{38.15} & 87.73 & 40.08 & 87.70 \\ 
 & ReAct\cite{sun2021react} & 17.77 & 96.70 & 25.15 & 94.34 & 34.64 & \textbf{91.92} & 51.31 & 88.83 & 32.22 & 92.95 \\ 
 & BATS(Ours) & \textbf{12.57} & \textbf{97.67} & \textbf{22.62} & \textbf{95.33} & \textbf{34.34} & 91.83 & 38.90 & \textbf{92.27} & \textbf{27.11} & \textbf{94.28} \\ \hline 
\multirow{7}{*}{DN121} & MSP\cite{hendrycks17baseline} & 47.65 & 89.09 & 69.95 & 79.64 & 72.53 & 78.74 & 69.69 & 77.06 & 64.96 & 81.13 \\ 
 & ODIN\cite{ODIN} & 30.72 & 93.66 & 57.90 & 86.11 & 63.16 & 83.54 & 53.51 & 83.88 & 51.32 & 86.80 \\ 
 & Energy\cite{liu2020energy} & 33.16 & 93.81 & 53.79 & 86.70 & 61.01 & 83.83 & 55.42 & 84.06 & 50.85 & 87.10 \\ 
 & Mahalanobis\cite{Mahalanobis} & 97.36 & 42.24 & 96.21 & 41.28 & 97.32 & 47.27 & 62.78 & 56.53 & 88.42 & 46.83 \\ 
 & GradNorm\cite{huang2021importance} & 22.88 & 94.40 & 43.12 & 87.55 & 55.80 & 82.00 & 47.58 & 85.16 & 42.35 & 87.28 \\  
 & ReAct\cite{sun2021react} & 15.93 & 96.91 & 40.41 & 90.13 & 48.87 & 87.98 & 36.58 & 92.48 & 35.45 & 91.88 \\ 
 & BATS(Ours) & \textbf{14.63} & \textbf{97.13} & \textbf{30.45} & \textbf{93.03} & \textbf{41.35} & \textbf{89.24} & \textbf{31.72} & \textbf{93.40} & \textbf{29.54} & \textbf{93.20} \\ \hline
\multirow{7}{*}{MNet} & MSP\cite{hendrycks17baseline} & 63.09 & 85.71 & 79.67 & 76.01 & 81.47 & 75.51 & 75.12 & 76.49 & 74.84 & 78.43 \\
 & ODIN\cite{ODIN} & 45.61 & 91.33 & 63.03 & 83.44 & 70.01 & 80.85 & 52.45 & 85.61 & 57.78 & 85.31 \\
 & Energy\cite{liu2020energy} & 49.52 & 91.10 & 63.06 & 84.42 & 69.24 & 81.42 & 58.16 & 84.88 & 60.00 & 85.46 \\
 & Mahalanobis\cite{Mahalanobis} & 62.04 & 82.37 & 54.79 & 86.33 & 53.77 & 83.69 & 88.72 & 37.28 & 64.83 & 72.42 \\
 & GradNorm\cite{huang2021importance} & 33.70 & 92.46 & 42.15 & 89.65 & 56.56 & 83.93 & \textbf{34.95} & \textbf{90.99} & 41.84 & 89.26 \\
 & ReAct\cite{sun2021react} & 37.08 & 93.41 & 53.13 & 86.04 & 54.15 & 83.31 & 42.45 & 89.42 & 46.70 & 88.05 \\
 & BATS(Ours) & \textbf{31.56} & \textbf{94.33} & \textbf{41.68} & \textbf{90.21} & \textbf{52.43} & \textbf{86.26} & 38.69 & 90.76 & \textbf{41.09} & \textbf{90.39} \\ \hline 
\end{tabular}\label{tab:imagenet}}
\end{table}

\subsection{Evaluation on the large-scale OOD detection benchmark}
We first evaluate our method on a large-scale OOD detection benchmark proposed by \citet{huang2021mos}. \cite{huang2021mos} revealed that OOD detection methods designed for the CIFAR benchmark might not effectively be adaptable for the ImageNet benchmark with a large semantic space.
Recent {literature} \cite{sun2021react,huang2021importance,huang2021mos} proposes to evaluate OOD detection performance on images that have higher resolution and contain more classes than the CIFAR benchmarks, which is more relevant to real-world applications.

In Tab. \ref{tab:imagenet}, we compare our method with the existing methods and show the OOD detection
performance for each OOD test dataset and the average over the four datasets. We consider different architectures, including the widely used ResNet-50 \cite{he2016resnet}, DenseNet-121 \cite{huang2018densenet} and a lightweight model MobileNet-v2 \cite{sandler2019mobilenetv2}. Compared with the Energy Score \cite{liu2020energy}, the difference in our approach is rectifying the features that deviate from the feature's typical set. Our method outperforms the Energy Score on ResNet-50 by 31.05$\%$ in FPR95 and 8.46$\%$ in AUROC. Furthermore, our method reduces FPR95 by 5.11$\%$ and improves AUROC by 1.33$\%$ compared to the state-of-the-art method \cite{sun2021react} on ResNet-50. 
{Here the models are pre-trained in a standard manner.}
We also show that \textbf{BATS} can boost the OOD detection when using the adversarially pre-trained classifiers in Appendix \ref{App:Robust}.

Simultaneously, we observe that existing methods have different performances on different architectures. {Specifically}, the performance of the GradNorm \cite{huang2021importance} in FPR95 is 7.86$\%$ worse than that of ReAct \cite{sun2021react} on the ResNet-50, but surpasses ReAct on MobileNet-V2 by 4.86$\%$. Our method achieves the best performance on different architectures.
Appendix \ref{App:NAE} shows that our method also outperforms the existing methods when choosing the natural adversarial examples \cite{hendrycks2021nae} as OOD examples.

 \begin{figure}[htbp]
\centering
\includegraphics[width=0.80\textwidth]{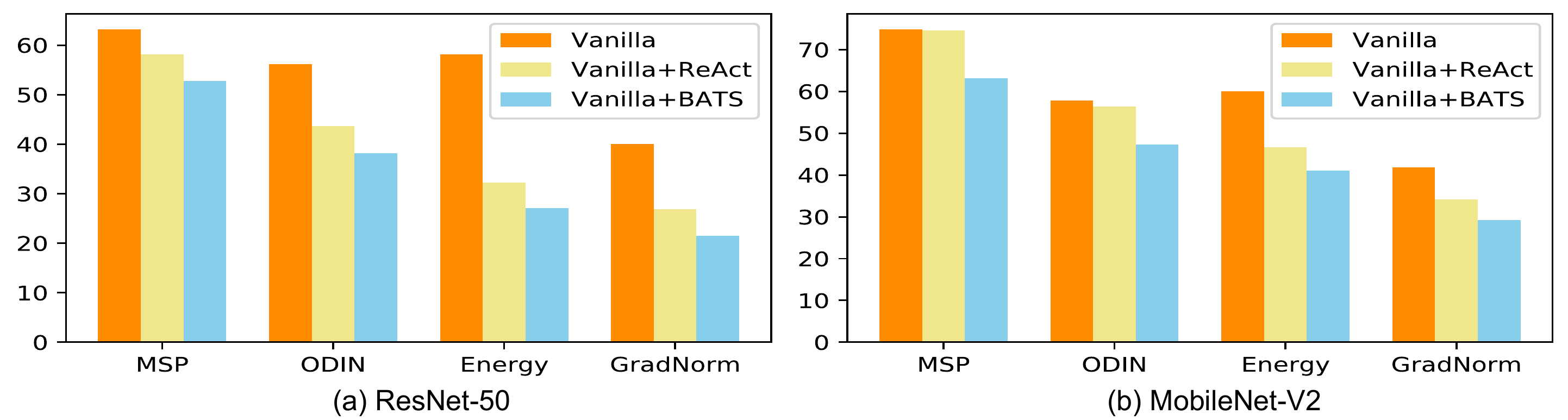}
\caption{The FPR95 for different methods on ImageNet (lower is better) on ResNet-50 and MobileNet-V2. We illustrate the average performance on four OOD datasets. "Vanilla" means the original method and "Vanilla+BATS" means applying our \textbf{BATS} on the method.}
\label{img:combine}
\end{figure}

Our experiments mainly use the energy score as the test statistic. In Fig. \ref{img:combine}, we show that \textbf{BATS} is also compatible with various OOD scores and \textbf{BATS} can boost the performance of various OOD scores. Applying our \textbf{BATS} on GradNorm \cite{huang2021importance} (a gradient-based OOD score) can even achieve better performance than "Energy+BATS" but this method needs to derive the gradients of the model, which costs more than "Energy+BATS." See Appendix \ref{App:combine} for detailed performance.

 

\begin{table}[htbp]
\centering
\caption{OOD detection performance on CIFAR-10 and CIFAR-100 \cite{krizhevsky2009learningCIFAR}. All methods are post hoc and can be directly used for pre-trained models. The best results are in Bold.}
\scalebox{0.712}{
\begin{tabular}{cccccccccccc}
\hline 
\multirow{3}{*}{Dataset} & \multirow{3}{*}{Method} & \multicolumn{2}{c}{SVHN} & \multicolumn{2}{c}{Tiny-Imagenet} & \multicolumn{2}{c}{LSUN\_resize} & \multicolumn{2}{c}{Texture} & \multicolumn{2}{c}{Average} \\ \cline{3-12} 
 &  & FPR95 & AUROC & FPR95 & AUROC & FPR95 & AUROC & FPR95 & AUROC & FPR95 & AUROC \\
 &  & $\downarrow$ & $\uparrow$ & $\downarrow$ & $\uparrow$ & $\downarrow$ & $\uparrow$ & $\downarrow$ & $\uparrow$ & $\downarrow$ & $\uparrow$\\ \hline
\multirow{6}{*}{\makecell[c]{CIFAR10\\RN18}} & MSP\cite{hendrycks17baseline} & 59.60 & 91.29 & 50.01 & 93.02 & 52.15 & 92.73 & 66.63 & 88.50 & 57.10 & 91.39 \\
 & ODIN\cite{ODIN} & 59.71 & 88.52 & \textbf{10.95} & \textbf{98.08} & \textbf{9.24} & \textbf{98.25} & 52.06 & 89.16 & 32.99 & 93.50 \\
 & Energy\cite{liu2020energy} & 54.03 & 91.32 & 15.18 & 97.28 & 23.53 & 96.14 & 55.30 & 89.37 & 37.01 & 93.53 \\
 & GradNorm\cite{huang2021importance} & 82.45 & 79.85 & 19.23 & 96.77 & 48.99 & 90.67 & 69.40 & 81.72 & 55.02 & 87.25 \\
 & ReAct\cite{sun2021react} & 46.87 & 92.54 & 22.80 & 96.10 & 18.31 & 96.92 & 47.39 & 91.58 & 33.84 & 94.29 \\
 & BATS(Ours) & \textbf{38.42} & \textbf{93.53} & 17.75 & 96.91 & 19.85 & 96.59 & \textbf{43.81} & \textbf{92.32} & \textbf{29.96} & \textbf{94.84} \\ \hline 
 
 \multirow{6}{*}{\makecell[c]{CIFAR10\\WRN}} & MSP\cite{hendrycks17baseline} & 63.24 & 86.66 & 39.57 & 94.60 & 44.31 & 93.82 & 60.71 & 88.90 & 51.96 & 91.00 \\
 & ODIN\cite{ODIN} & 61.13 & 82.49 & \textbf{12.79} & \textbf{97.61} & 12.49 & 97.50 & 61.13 & 80.18 & 36.89 & 89.45 \\
 & Energy\cite{liu2020energy} & 56.05 & 86.63 & 17.58 & 96.99 & 28.44 & 95.29 & 61.74 & 85.68 & 40.95 & 91.15 \\
 & GradNorm\cite{huang2021importance} & 88.55 & 49.14 & 41.25 & 90.68 & 91.02 & 48.94 & 90.83 & 46.28 & 77.91 & 58.76 \\
 & ReAct\cite{sun2021react} & 58.35 & 86.67 & 18.85 & 96.62 & 16.52 & 97.04 & 50.89 & 89.27 & 36.15 & 92.40 \\
 & BATS(Ours) & \textbf{50.60} & \textbf{89.50} & 25.17 & 95.66 & \textbf{11.98} & \textbf{97.70} & \textbf{45.30} & \textbf{91.18} & \textbf{33.26} & \textbf{93.51} \\ \hline 
 
\multirow{6}{*}{\makecell[c]{CIFAR100\\RN18}} & MSP\cite{hendrycks17baseline} & 81.79 & 77.80 & 68.32 & 83.92 & 82.51 & 75.73 & 85.12 & 73.36 & 79.44 & 77.70 \\
 & ODIN\cite{ODIN} & \textbf{40.82} & \textbf{93.32} & 69.34 & 86.28 & 79.62 & 82.12 & 83.61 & 72.36 & 68.35 & 83.52 \\
 & Energy\cite{liu2020energy} & 81.24 & 84.59 & 40.12 & 93.16 & 73.56 & 82.98 & 85.87 & 74.94 & 70.20 & 83.92 \\
 & GradNorm\cite{huang2021importance} & 57.65 & 87.77 & \textbf{25.77} & \textbf{95.12} & 89.60 & 63.25 & 79.08 & 68.89 & 63.03 & 78.76 \\
 & ReAct\cite{sun2021react} & 70.28 & 88.25 & 45.62 & 91.02 & 55.57 & 89.32 & 61.01 & 87.57 & 58.12 & 89.04 \\
 & BATS(Ours) & 61.48 & 90.63 & 44.41 & 91.27 & \textbf{52.68} & \textbf{90.04} & \textbf{52.36} & \textbf{89.72} & \textbf{52.73} & \textbf{90.42} \\ \hline 
 
 \multirow{6}{*}{\makecell[c]{CIFAR100\\WRN}} & MSP\cite{hendrycks17baseline} & 78.43 & 77.74 & 61.33 & 87.46 & 81.69 & 72.69 & 85.07 & 75.46 & 76.63 & 78.34 \\
 & ODIN\cite{ODIN} & \textbf{35.69} & \textbf{94.84} & 82.68 & 79.17 & 87.48 & 74.53 & 86.97 & 65.40 & 73.21 & 78.49 \\
 & Energy\cite{liu2020energy} & 75.57 & 83.05 & \textbf{40.87} & \textbf{92.99} & 65.90 & 82.78 & 87.98 & 71.21 & 67.58 & 82.51 \\
 & GradNorm\cite{huang2021importance} & 83.24 & 72.55 & 45.20 & 90.43 & 78.62 & 68.80 & 92.59 & 46.99 & 74.91 & 69.69 \\
 & ReAct\cite{sun2021react} & 72.94 & 86.89 & 42.07 & 91.97 & 60.87 & 85.90 & 84.18 & 76.22 & 65.02 & 85.25 \\
 & BATS(Ours) & 71.01 & 87.50 & 41.93 & 91.97 & \textbf{57.01} & \textbf{88.04} & \textbf{80.46} & \textbf{78.42} & \textbf{62.60} & \textbf{86.48} \\ \hline 
\end{tabular}\label{tab:CIFAR}}
\end{table}

\subsection{Evaluation on CIFAR benchmarks}

We further evaluate our method on CIFAR benchmarks and use CIFAR-10 and CIFAR-100 \cite{krizhevsky2009learningCIFAR} as the in-distribution datasets respectively. Tab. \ref{tab:CIFAR} compares our method with the baseline methods and shows the OOD detection
performance for each OOD test dataset and the average over the four datasets. We evaluate our method on the ResNet-18 (RN18) \cite{he2016resnet} and WideResNet-28-10 (WRN) \cite{zagoruyko2017wideresnet}. The models are trained for 200 epochs with a batch size of 128. The {starting} learning rate is 0.1 and decays by a factor of 10 at epochs 100 and 150.

ODIN \cite{ODIN} performs the best in the baselines methods on CIFAR-10 with an FPR95 of 32.99$\%$ on ResNet-18. Our method outperforms ODIN by 3.03$\%$ and outperforms the simple baseline method MSP \cite{hendrycks17baseline} by 27.14$\%$ in FPR95. As for using CIFAR-100 as the in-distribution dataset, ReAct \cite{sun2021react} is the best baseline method. Our approach surpasses the ReAct by 5.39$\%$ in FPR95 on ResNet-18. Our method achieves the best performance on both CIFAR-10 and CIFAR-100. Our approach is also effective when using the WideResNet model, outperforming the existing methods.


\subsection{Ablation studies}
\subsubsection{Rectifying the features of the early layers} 
In our experiments, we rectify the features of the penultimate layer (the layer before the fully connected layer), which is convenient and efficient. However, \textbf{what will happen if we rectify the features of the early layers with BATS?} The early layers refer to the layers close to the input \cite{natureindividual}. In particular, the original ResNet-50 \cite{he2016resnet} consists of four residual blocks. Block1 is close to the input and Block4 is close to the output. In Tab. \ref{tab:blocks}, we show the influence of applying feature rectification on the output of different blocks. Applying feature rectification to the early blocks (from Block1 to Block3) has little effect on the performance of OOD detection, while the last block plays a vital role. Applying feature rectification on all the blocks performs the best in our experiments, which is 0.99$\%$ higher than the "Block4" in FPR95 and 32.04$\%$ higher than "Without" in FPR95. Considering that the latest block has a more significant impact on the OOD detection performance than the other blocks, we just rectify the features of the penultimate layer with \textbf{BATS} for the simplicity of the 
method. 

\begin{table}[htbp]
\centering
\caption{Ablation study of the influence of feature rectification on different blocks. "Without" means applying no rectification on any blocks. "Block1-4" means applying rectification on all blocks.} 
\scalebox{0.8}{
\begin{tabular}{ccccccccccc}
\hline 
\multirow{2}{*}{Blocks} & \multicolumn{2}{c}{iNaturalist} & \multicolumn{2}{c}{SUN} & \multicolumn{2}{c}{Places} & \multicolumn{2}{c}{Textures} & \multicolumn{2}{c}{Average} \\ \cline{2-11} 
 & FPR95 & AUROC & FPR95 & AUROC & FPR95 & AUROC & FPR95 & AUROC & FPR95 & AUROC \\ \hline
Without & 46.65 & 91.32 & 61.96 & 84.88 & 67.97 & 82.21 & 56.06 & 84.88 & 58.16 & 85.82 \\
Block1 & 49.72 & 90.73 & 62.67 & 84.62 & 68.30 & 82.03 & 55.62 & 85.04 & 59.08 & 85.61 \\
Block2 & 41.78 & 92.36 & 63.73 & 84.67 & 69.45 & 81.95 & 55.53 & 85.45 & 57.62 & 86.11 \\
Block3 & 40.76 & 92.55 & 58.37 & 86.56 & 64.78 & 83.82 & 51.45 & 86.77 & 53.84 & 87.43 \\
Block4 & \textbf{12.57} & \textbf{97.67} & 22.62 & 95.33 & 34.34 & 91.83 & 38.90 & 92.27 & 27.11 & 94.28 \\
Block1-2 & 43.63 & 91.92 & 63.22 & 84.61 & 69.28 & 81.84 & 53.67 & 85.72 & 57.45 & 86.02 \\
Block1-3 & 38.05 & 93.04 & 59.47 & 86.47 & 66.30 & 83.49 & 49.72 & 87.50 & 53.39 & 87.63 \\
Block1-4 & 12.76 & 97.54 & \textbf{21.15} & \textbf{95.51} & \textbf{33.01} & \textbf{91.91} & \textbf{37.55} & \textbf{92.54} & \textbf{26.12} & \textbf{94.38} \\ \hline 
\end{tabular}\label{tab:blocks}}
\end{table}

\begin{figure}[htbp]
\centering
\includegraphics[width=0.96\textwidth]{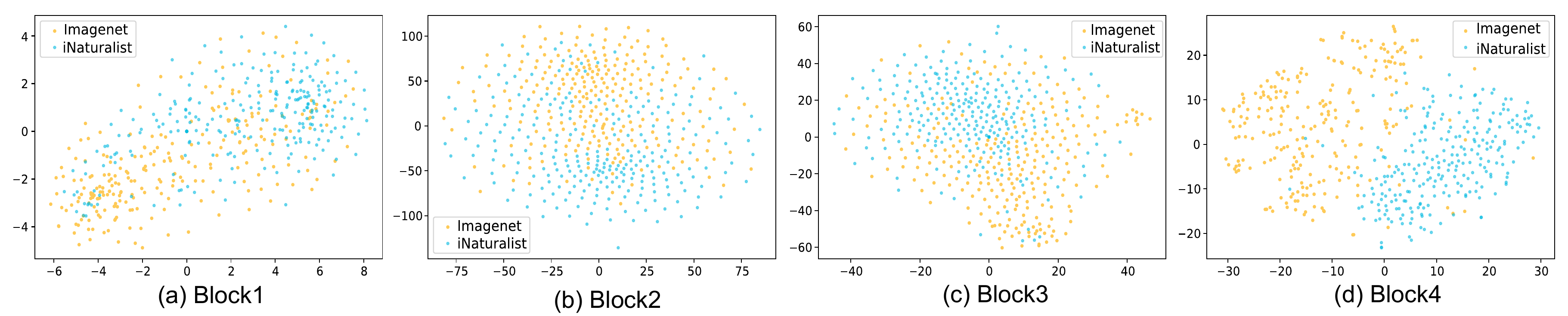}
\vspace{-0.4cm}
\caption{t-SNE visualizations. We illustrate the t-SNE plots for the features of in-distribution examples (ImageNet) and out-of-distribution examples (iNaturalist) from different blocks.}
\label{img:blocks_inaturalist}
\end{figure}

To find out why the last block has a significant influence on the OOD detection while the other blocks contribute little, we visualize the feature embeddings extracted by different blocks in ResNet-50 using t-SNE~\cite{tsne} in Fig. \ref{img:blocks_inaturalist}. We choose the iNaturalist as the OOD dataset and the ImageNet as the ID dataset. The features extracted by the early blocks of the ID and OOD examples are similar, which has little benefit in distinguishing the ID and OOD examples. In contrast, the last block can extract perfectly separable features for the ID and OOD examples. This may be due to the fact that deep neural networks focus on similar general features (edges, lines, and colors) in the early layers and pay more attention to specific features related to classification in the late layers \cite{natureindividual,yosinski2014transferable}. The late layer can contribute more to the OOD detection than the early layer. See more in Appendix \ref{App:early}. 

\subsubsection{The influence of the hyperparameter}\label{sec:influence}

\begin{figure}[htbp]
\centering
\includegraphics[width=0.75\textwidth]{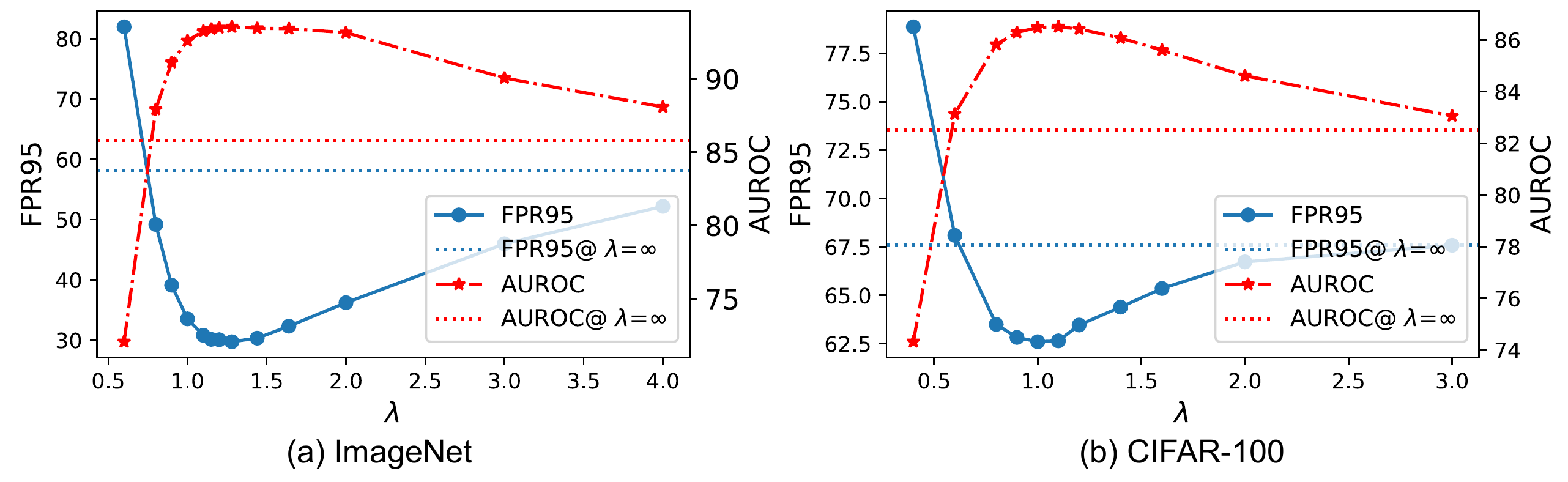}
\vspace{-0.3cm}
\caption{(a) The influence of the hyper-parameter $\lambda$ on the OOD detection on ImageNet. The model is ResNet-50. We illustrate the average performance on four OOD datasets. (b) The influence of the hyper-parameter $\lambda$ on the OOD detection on CIFAR-100. The model is WideResNet. The horizontal line indicates the OOD detection performance without feature rectification.}
\vspace{-0.3cm}
\label{img:lambda_lambda}
\end{figure}

In Sec. \ref{sec43}, we theoretically analyze the bias-variance trade-off in our method. 
{Our proposed \textbf{BATS} can reduce variance, which benefits OOD detection, but can also introduce a bias.}
Here, we empirically show the influence of the hyperparameter $\lambda$ in Fig. \ref{img:lambda_lambda}. As $\lambda$ tends to infinity, BATS approaches to the Energy Score (the horizontal lines). Very small $\lambda$ will damage the performance. 

\section{Conclusion}\label{Conclusion}

In this paper, we provide novel insight into the obstacle factor in OOD detection from the perspective of typicality and hypothesize that extreme features can be the culprit.
We propose to rectify the features into the typical set and provide a concise and effective post-hoc approach \textbf{BATS} to estimating the feature's typical set. \textbf{BATS} can be applied to various OOD scores to boost the OOD detection performance. Theoretical analysis and ablations provide a further understanding of our approach.
Experimental results show that our \textbf{BATS} can establish state-of-the-art OOD detection performance on the ImageNet benchmark, surpassing the previous best method by 5.11$\%$ in FPR and 1.43$\%$ in AUROC.
We hope that our findings can motivate new research into the internal mechanisms of deep models and OOD detection and uncertainty estimation from the perspective of feature typicality.

\textbf{Limitations and societal impact.} This paper proposes to rectify the feature into its typical set to improve the detection performance against OOD data and provides a plug-and-play method with the assistance of BN. The limitation of our method can be that the BN layers are required in the model architecture in our approach. BN layers are widely used in convolutional neural networks to alleviate covariate shifts, but there are also architectures without BN. A set of training images can contribute to selecting the feature's typical set and alleviate this limitation. We also anticipate some other information in the model is conducive to selecting the feature's typical set and improving the post-hoc OOD detection performance. We leave this as future work. {Although truncating features into a typical set can improve OOD detection, a potential negative impact of the proposed process is that it inherently introduces a bias and causes some information loss which may be important to the model in real-world scenarios.}

\section*{Acknowledgments}
This work was supported in part by the Fundamental Research Funds for the Central Universities, by Alibaba Group through Alibaba Research Intern Program, and by the National Nature Science Foundation of China 62001146. Dr. Xie's research work is partially supported by the Interdisciplinary Intelligence SuperComputer Center of Beijing Normal University at Zhuhai.



\bibliographystyle{unsrtnat}

\bibliography{reference}

\section*{Checklist}


\begin{enumerate}

\item For all authors...
\begin{enumerate}
  \item Do the main claims made in the abstract and introduction accurately reflect the paper's contributions and scope?
    \answerYes{See abstract and Section 1.}
  \item Did you describe the limitations of your work?
    \answerYes{We describe the limitation of our work in our Conclusion section.}
  \item Did you discuss any potential negative societal impacts of your work?
    \answerYes{We discuss the potential negative societal impacts of our work in the Conclusion section.}
  \item Have you read the ethics review guidelines and ensured that your paper conforms to them?
    \answerYes{}
\end{enumerate}

\item If you are including theoretical results...
\begin{enumerate}
  \item Did you state the full set of assumptions of all theoretical results?
    \answerYes{See Section 4.3. We theoretically analyze the bias-variance trade-off and state the full set of assumptions.}
        \item Did you include complete proofs of all theoretical results?
    \answerYes{See Appendix \ref{App:proof}}
\end{enumerate}

\item If you ran experiments...
\begin{enumerate}
  \item Did you include the code, data, and instructions needed to reproduce the main experimental results (either in the supplemental material or as a URL)?
    \answerNo{We introduce the implementation of our method in Section 5.1 and Appendix \ref{App:details}. We plan to open the source code to reproduce the main experimental results later.}
  \item Did you specify all the training details (e.g., data splits, hyperparameters, how they were chosen)?
    \answerYes{See the Section 5.1 and Appendix \ref{App:details} for details.}
        \item Did you report error bars (e.g., with respect to the random seed after running experiments multiple times)?
    \answerNA{}
        \item Did you include the total amount of compute and the type of resources used (e.g., type of GPUs, internal cluster, or cloud provider)?
    \answerYes{See Appendix \ref{App:details}.3}
\end{enumerate}

\item If you are using existing assets (e.g., code, data, models) or curating/releasing new assets...
\begin{enumerate}
  \item If your work uses existing assets, did you cite the creators?
    \answerYes{We use the pre-trained model in PyTorch and cite the creators.}
  \item Did you mention the license of the assets?
    \answerNA{}
  \item Did you include any new assets either in the supplemental material or as a URL?
    \answerNA{}
  \item Did you discuss whether and how consent was obtained from people whose data you're using/curating?
    \answerNA{}
  \item Did you discuss whether the data you are using/curating contains personally identifiable information or offensive content?
    \answerNA{}
\end{enumerate}

\item If you used crowdsourcing or conducted research with human subjects...
\begin{enumerate}
  \item Did you include the full text of instructions given to participants and screenshots, if applicable?
    \answerNA{}
  \item Did you describe any potential participant risks, with links to Institutional Review Board (IRB) approvals, if applicable?
    \answerNA{}
  \item Did you include the estimated hourly wage paid to participants and the total amount spent on participant compensation?
    \answerNA{}
\end{enumerate}

\end{enumerate}


\appendix


\section{Type I error and type II error in OOD detection}\label{App:TypeError}
In the preliminary section, we provide a summary for the out-of-distribution detection from the perspective of hypothesis testing. As for the error of an OOD detection method, it can be evaluated from two dimensions.
The mistaken rejection of an actually true null hypothesis $\cH_0$ is the type I error. 
The significance level $\alpha$ is a predetermined scalar that bounds the type I error above:
\[
\alpha >= P( \rvx \in \cR | \cH_0) = P(T(\rvx;f) \geq \gamma | \cH_0)= P_0(T(\rvx;f) \geq \gamma).
\]
By the Neyman–Pearson lemma \cite{neyman1933ix}, the threshold $\gamma$ is determined by solving the equation $\alpha = P_0(T(\rvx;f) \geq \gamma).$
For a given significance level, the goal is to minimize the type II error: the failure to reject a null hypothesis that is actually false. The probability of the type II error is denoted by
\[
\beta = P( x \notin \cR | \cH_1) = P(T(\rvx;f) < \gamma | \cH_1).
\]
In the literature on OOD detection, the type II error is also denoted by ``FPR$(1-\alpha)$", which is short for ``the false positive rate of OOD examples when the true positive rate for ID examples is $(1-\alpha)\%.$" In the experiments, we follow the notation FPR$(1-\alpha)$. 

\begin{figure}[htbp]
\centering
\includegraphics[width=0.85\textwidth]{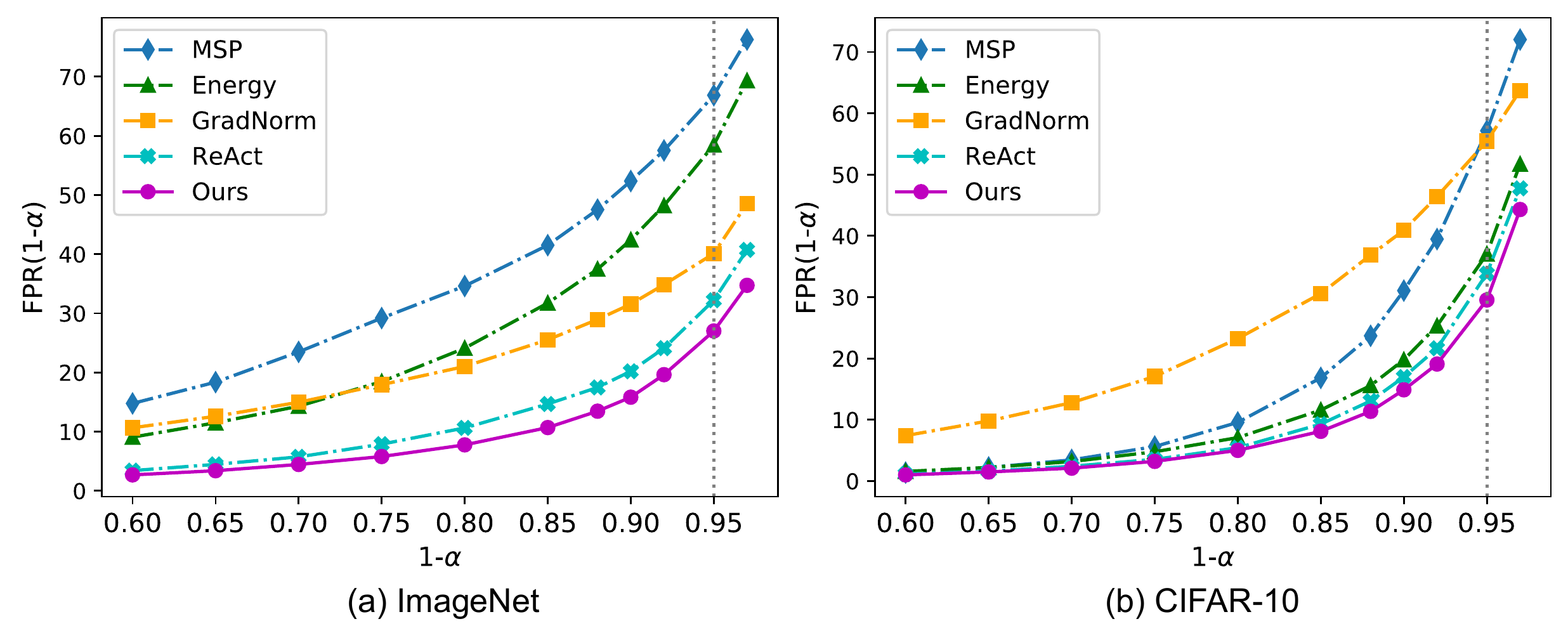}
\caption{(a) The FPR(1-$\alpha$) for different methods on ImageNet (lower is better). The model is ResNet-50. We illustrate the average performance on four OOD datasets. The grey vertical line indicates the performance in FPR95. (b) The FPR(1-$\alpha$) for different methods on CIFAR-10. The model is ResNet-18.}
\label{img:alpha}
\end{figure}

In our paper, we mainly show the superiority of our method on different datasets in the metrics of FPR95 and AUROC. Here we illustrate the OOD detection performance at different significance levels (FPR(1-$\alpha$)) in Fig. \ref{img:alpha}. The horizontal axis represents the significance level for FPR ("0.95" means FPR95). Our method surpasses the existing methods at different significance levels on both the large scale dataset (ImageNet) and the small scale dataset (CIFAR-10).

\section{The influence of the early layers}\label{App:early}
In the paper, we show that the early layers of the model can hardly distinguish the feature embeddings of the in-distribution examples (ImageNet-1k) and out-of-distribution examples (iNaturalist) for that the features of these examples extracted by the early layers are mixed up. Rectifying the features of the early layers contributes little to OOD detection.
In this section, we illustrate the t-SNE visualization for the feature embeddings of in-distribution examples and other out-of-distribution examples (Places \cite{zhou2017places}, SUN \cite{xiao2010sun}, and Textures \cite{cimpoi2014Texture}) from different blocks. Their t-SNE visualization results are similar. To be specific, the feature embeddings of the early blocks of the different datasets are similar, while the last block shows differences. In Tab. \ref{tab:blocks}, we set the $\lambda$ for Block1 and Block2 as 5, the $\lambda$ for Block3 as 2, and the $\lambda$ for Block4 as 1, for that restricting the features of the early layers may have a negative impact on the late layers.

\begin{figure}[htbp]
\centering
\includegraphics[width=1.0\textwidth]{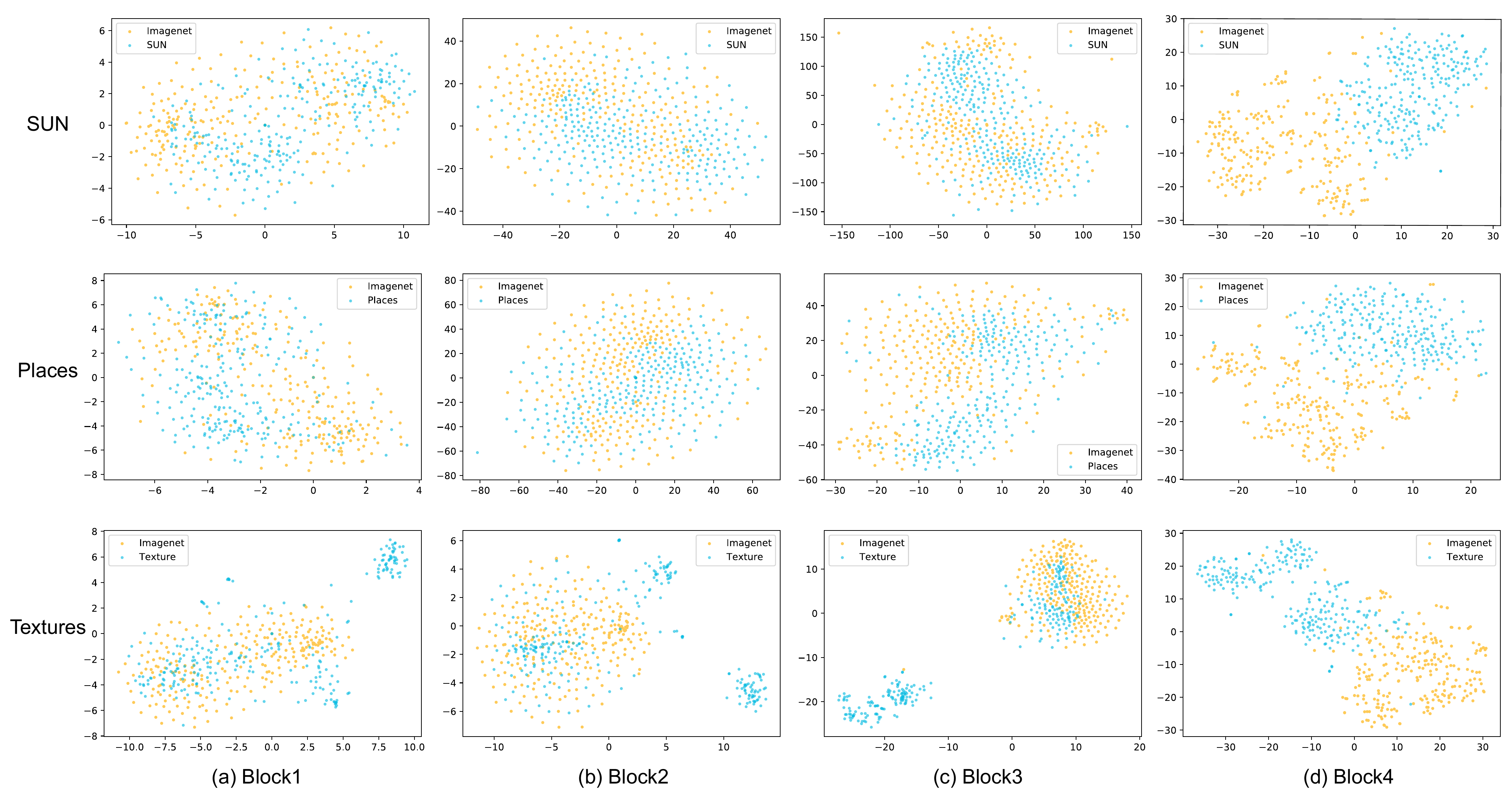}
\caption{t-SNE visualization for the feature embeddings of in-distribution examples and out-of-distribution examples from different blocks. The model we used is ResNet-50.}
\label{img:other_blocks}
\end{figure}

\section{Energy and density in the classifier}\label{App:energy}

 To make our paper self-contained, we provide some details for the energy and density in the classifier with reference to the previous works \cite{energyrobust,grathwohl2019your,liu2020energy}.

\citet{lecun2006} show that any probability density $p(x)$ for $x$ can be expressed as
\begin{equation}
    p(x) = \frac{\exp(-E(x))}{Z}, 
\label{px0_}
\end{equation}   
where $E(x)$ represents the energy of $x$ and is modeled by neural network, $Z = \int\exp(-E(x))dx$ is the normalizing factor which is also known as the partition function.

Similarly, $p(x,y)$ can be defined as follows:
\begin{equation}
    p(x,y) =  \frac{\exp(-E(x,y))}{\tilde{Z}},
\label{pxy0_}
\end{equation}
where $\tilde{Z} = \int\sum\limits_{y}\exp(-E(x,y))dx$.

Thus we also get $p(y|x)$ expressed by $E(x)$ and $E(x,y)$:
\begin{equation}
    p(y|x) = \frac{p(x,y)}{p(x)} = \frac{\exp(-E(x,y)) \cdot Z}{\exp(-E(x)) \cdot \tilde{Z}}.
\label{pxgiveny0_}
\end{equation}

We denote $f$ as a classification neural network. Let $x$ be a sample. Then $f(x)[k]$ represents the $k^{th}$ output of the last layer and $p(y|x)$ can be defined as:
\begin{equation}
p(y|x) = \frac{\exp({f(x)[y]})}{\sum_{k=1}^{n} \exp({f(x)[k]})},
\label{p(ygx_)}    
\end{equation}
where $n$ represents total possible classes. 
From Eq. (\ref{pxgiveny0_}) and (\ref{p(ygx_)}), we define two energy functions as follows:
\begin{equation}
\left\{
\begin{array}{l}
E(x,y) = -\log(\exp({f(x)[y]})), \\
E(x) = -\log(\sum_{k=1}^{n} \exp({f(x)[k]})). 
\end{array}
\right.
\label{energy_x}
\end{equation}
And thus $Z$ can be expressed as:
\begin{equation}
\begin{array}{ll}
Z & = \int_{x}\exp(-E(x))dx = \int_{x}\exp(\log(\sum\limits_{y} \exp({f(x)[y]})))dx = \int_{x}(\sum\limits_{y} \exp({f(x)[y]}))dx    = \tilde{Z}.       
\end{array}  
\label{ztheta}
\end{equation}

From Eq. (\ref{energy_x}) and Eq. (\ref{px0_}), the marginal density $p(x)$ for $x$ can be expressed by the output of the classifier as:
\begin{equation}
    p(x) = \frac{\sum_{k=1}^{n} \exp({f(x)[k]})}{Z},
\label{px0__}
\end{equation}   
where $Z$ is independent to $\rvx$.

\section{Proofs of section~\ref{sec43}}\label{App:proof}

In this section, we prove the main results in Section~\ref{sec43}.
Recall the layer structure in (\ref{layer}):
\benr
\rvz' \rarrow \text{BN}(\rvz'; \mu, \sigma)\text{ or } \text{TrBN}(\rvz'; \mu, \sigma,\lambda)\rarrow \text{ReLU}  \rarrow \rvz,
\eenr
where $\rvz'$ is the feature vector extracted from the penultimate layer of $g.$ We denote 
\benr
\rvz_1 = \text{BN}(\rvz'; \mu, \sigma) \quad \text{and} \quad \bar \rvz_1 = \text{TrBN}(\rvz'; \mu, \sigma). 
\eenr
Suppose $\rvz'$ is a Gaussian variable. Then $\rvz_1 \sim N(\mu, \sigma^2)$ and $\bar \rvz_1$ follows a Rectified Gaussian distribution with lower bound $\mu-\lambda\sigma$ and upper bound $\mu+\lambda\sigma.$ 
The cdf of $\bar \rvz_1$ is
\benr
F^R(\rvz'| \mu, \sigma^2) =  \begin{cases}
0, & \text{if} \quad \rvz' <a ; \\
\Phi(\rvz'; \mu, \sigma^2),  & \text{if}  \quad a \leq \rvz' < b; \\
1, & \text{if}\quad b \leq \rvz',
\end{cases}
\eenr
where $\Phi(\rvz'; \mu, \sigma^2)$ the cdf of a normal distribution with mean $\mu$ and variance $\sigma^2.$
According to \cite{PALMER201751}, 
\benr
\E(\bar \rvz_1) = \mu \quad \text{and} \quad \text{Var}(\bar \rvz_1) = \sigma^2 C(\lambda),
\eenr
where
\benr
C(\lambda) =  \text{erf}(\frac{\lambda}{\sqrt 2}) - \frac{\sqrt 2}{\sqrt \pi}\lambda \exp(-\frac{\lambda^2}{2}) + \lambda^2(1-\text{erf}(\frac{\lambda}{\sqrt 2})),
\eenr
and $\text{erf}(x) = (2/\sqrt \pi) \int_0^x \exp(-t^2) dt$ is the Gauss error function.
It is easy to see
\benr
C(0)=0 \quad \text{and} \quad C'(\lambda) = 2 \lambda \cdot \text{erf}(\frac{\lambda}{\sqrt  2}) > 0.
\eenr
In addition,
\benr
\lambda^2(1-\text{erf}(\frac{\lambda}{\sqrt 2})) &=&  \lambda^2 \frac{2}{\sqrt \pi} \int_{\lambda/\sqrt 2}^{+\infty} \exp(-t^2) \mathrm{d} t \\
&\leq& \frac{4}{\sqrt \pi} \int_{\lambda/\sqrt 2}^{+\infty} t^2 \exp(-t^2) \mathrm{d} t \to 0, \quad \lambda \to +\infty. \nonumber
\eenr
Therefore, as $\lambda$ tends to $+\infty$,
\benr
\text{erf}(\frac{\lambda}{\sqrt 2}) \to 1, \quad \frac{\sqrt 2}{\sqrt \pi}\lambda \exp(-\frac{\lambda^2}{2}) \to 0, \quad \lambda^2(1-\text{erf}(\frac{\lambda}{\sqrt 2}))\to 0.
\eenr
We obtain that $C(\lambda) \to 1$ as $\lambda\to +\infty.$

Next we deal with the bias term. To proceed further, we need more notations as follow:
\benr
\rvz = \text{ReLU}(\text{BN}(\rvz'; \mu, \sigma)) \quad \text{and} \quad \bar \rvz = \text{ReLU}(\text{TrBN}(\rvz'; \mu, \sigma)). 
\eenr
Then we know that $\rvz$ follows a Rectified Gaussian distribution with lower bound $0$ and upper bound $+\infty$
and $\bar \rvz$ is a Rectified Gaussian variable with lower bound $0$ and upper bound $\mu+ \lambda \sigma.$
According to \cite{PALMER201751}, their expectations are
\benr
\E(\rvz) = \mu + \sigma \Big(\frac{1}{\sqrt{2\pi}} \big( \exp(-\frac{\mu^2}{2\sigma^2})  \big)   - \frac{\mu}{2\sigma}(1+\text{erf}(-\frac{\mu}{\sqrt{2}\sigma})) \Big),
\eenr
and
\benr
\E(\bar \rvz) &=& \mu + \sigma \Big(\frac{1}{\sqrt{2\pi}} \big( \exp(-\frac{\mu^2}{2\sigma^2}) - \exp(-\frac{\lambda^2}{2})  \big)   \\
&& - \frac{\mu}{2\sigma}(1+\text{erf}(-\frac{\mu}{\sqrt{2}\sigma})) + \frac{\lambda}{2} (1-\text{erf}(\frac{\lambda}{\sqrt 2}))\Big). \nonumber
\eenr
Therefore the bias term is
\benr
\E(\bar \rvz)-\E(\rvz) = \sigma\Big( - \exp(-\frac{\lambda^2}{2}) + \frac{\lambda}{2} (1-\text{erf}(\frac{\lambda}{\sqrt 2})) \Big).
\eenr
In addition,
\benr
\frac{\lambda}{2} (1-\text{erf}(\frac{\lambda}{\sqrt 2})) &=& \frac{\lambda}{2}\frac{2}{\sqrt \pi} \int_{\lambda/\sqrt{2}}^{+\infty} \exp(-t^2) \mathrm{d}t \\
&\leq& \frac{\sqrt 2}{\sqrt \pi} \int_{\lambda/\sqrt{2}}^{+\infty} t \exp(-t^2) \mathrm{d}t \to 0 \quad \text{as} \quad \lambda \to +\infty. \nonumber
\eenr
Then we obtain that
\benr
\text{Bias} = \E(\bar \rvz)-\E(\rvz) \to 0 \quad \text{as} \quad \lambda\to +\infty. 
\eenr

\section{Experiments details}\label{App:details}
\subsection{Details for metrics}
\textbf{FPR95:} the false positive rate of OOD (negative) examples when the true positive rate of in-distribution (positive) examples is as high as 95$\%$. The \textbf{t}rue \textbf{p}ositive \textbf{r}ate (TPR) can be computed as:
\begin{equation}
    TPR = \frac{TP}{(TP+FN)},
\label{Eq:TPR}
\end{equation}
where TP denotes the true positive (correctly identify the in-distribution examples as in-distribution examples) and FN denotes the False Negative (incorrectly identity the in-distribution examples as out-of-distribution examples). The \textbf{f}alse \textbf{p}ositive \textbf{r}ate (FPR) can be computed as:
\begin{equation}
    FPR = \frac{FP}{(FP+TN)},
\label{Eq:FPR}
\end{equation}
where FP denotes the false positive (incorrectly identify the out-of-distribution examples as in-distribution examples) and TN denotes the true negative (correctly identify the out-of-distribution examples as out-of-distribution examples).

\textbf{AUROC:} the area under the receiver operating characteristic curve (ROC) which is the plot of
TPR vs FPR. If FPR = 0 and TPR = 1, it means that this is a perfect OOD detector, which identify all examples correctly. If FPR=1 and TPR=0, this is a terrible detector that can not make any correct prediction. The closer the area under the ROC curve is to 1, the better the performance of the detector.

\subsection{Details for datasets}

\subsubsection{CIFAR OOD detection}

We use the CIFAR-10 and CIFAR-100 \cite{krizhevsky2009learningCIFAR} as the in-distribution examples respectively. CIFAR-10 consists of 60,000 images in the shape of $3 \times 32 \times 32$, including 10 categories (aircraft, cars, birds, cats, deer, dogs, frogs, horses, boats, and trucks). CIFAR-100 contains 100 categories of images, and each category has 600 images in the shape of $3 \times 32 \times 32$. We evaluate our approach on four common OOD datasets. We set $\lambda=3$ in our approach for CIFAR-10 and $\lambda=1.5$ and for CIFAR-100 on ResNet-18. As for WideResNet-28-10, we set $\lambda=0.7$ for CIFAR-10 and $\lambda=1.0$ for CIFAR-100. During the evaluation, all images are resized to $3 \times 32 \times 32$.

\textbf{SVHN} \cite{netzer2011readingSVHN}: Street View House Number (SVHN) consists of the house numbers extracted from Google Street View images. We use the entire of its test set as OOD examples (26032 images).


\textbf{Tiny ImageNet} \cite{chrabaszcz2017downsampledTiny}: Similar to ImageNet, Tiny ImageNet is an image classification dataset, which contains 200 categories, and each category contains 50 test images. We randomly crop the images to $3 \times 32 \times 32$.

\textbf{LSUN} \cite{yu2015lsun}: LSUN is a scene understanding dataset, which mainly includes scene images of bedrooms, fixed houses, living rooms, classrooms, etc. We randomly sample 10000 images as out-of-distribution examples and resize the images to $3 \times 32 \times 32$.

\textbf{Textures} \cite{cimpoi2014Texture}: Describable Textures Dataset (DTD) is a texture dataset, including 5640 images, which can be divided into 47 categories according to human perception. We use the entire Textures dataset for evaluation.

\subsubsection{Large-scale OOD detection}

We use the subsets from the following datasets as OOD examples and follow the setting in \cite{sun2021react} and \cite{huang2021importance}. The subsets are curated to be disjoint from the ImageNet-1k labels. We set $\lambda=1.25$ in our approach for ResNet-50 and DenseNet-121, and $\lambda=0.4$ for MobileNet-V2. During the evaluation, all images are resized to $3 \times 224 \times 224$. All models here use the softplus non-linearity ($\beta$ = 35), which can be expressed as the expectation of ReLU in a neighborhood \cite{zhu2022rethinking} and provide more robust features.

\textbf{iNaturalist} \cite{van2018inaturalist}: iNaturalist contains 675,170 training and validation images from 5089 natural fine-grained categories, including 13 major categories such as plants, insects, birds, and mammals. We randomly sample 10000 images that are disjoint from ImageNet-1k for evaluation.

\textbf{Places} \cite{zhou2017places}: Places is a scene image dataset, which contains 10 million pictures and more than 400 different types of scene environments. We randomly sample 10000 images that are disjoint from ImageNet-1k for evaluation.

\textbf{SUN} \cite{xiao2010sun}: The Scene UNderstanding (SUN) contains 397 well-sampled categories to evaluate the performance of scene recognition algorithms. We randomly sample 10000 images that are disjoint from ImageNet-1k for evaluation.

\textbf{Textures} \cite{cimpoi2014Texture}: Describable Textures Dataset (DTD) is a texture dataset, including 5640 images, which can be divided into 47 categories according to human perception. We use the
entire Textures dataset for evaluation.

\subsection{Hardware}
Our experiments are implemented by PyTorch \cite{pytorch} and runs on RTX-2080TI.

\section{Boosting the OOD detection on the robust classifiers}\label{App:Robust}
\begin{figure}[htbp]
\centering
\includegraphics[width=0.98\textwidth]{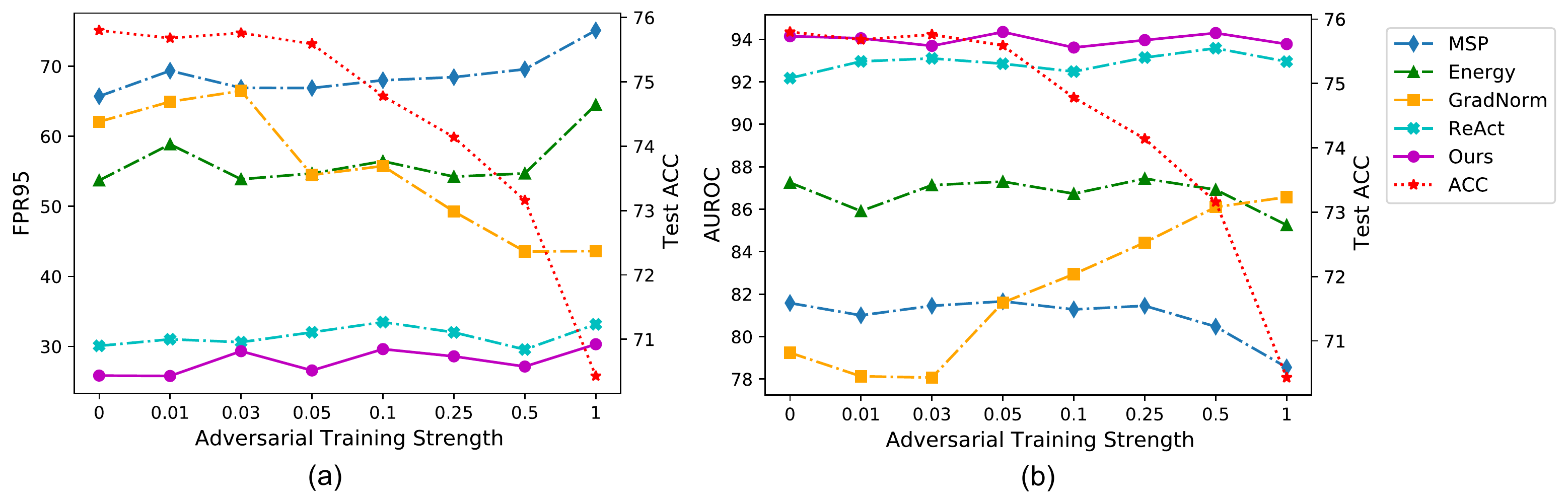}
\caption{(a) The FPR95 for different methods on ImageNet benchmark (lower is better) on different robust models. We illustrate the average performance on four OOD datasets. The red dotted line indicates the test accuracy of the robust model. (b) The AUROC for different methods (higher is better).}
\label{img:advmodel}
\end{figure}

 The models used in our experiments (Tab. \ref{tab:imagenet}) are standard pre-trained. \citet{salman2020adversariallytrans} show that adversarially robust models with less accuracy often perform better than their standard-trained counterparts in transfer learning.
 In Fig. \ref{img:advmodel} we evaluate the OOD detection performance on eight adversarially pre-trained ResNet-50 trained with $\ell_{2}$ perturbation of different strength $\epsilon$ \cite{salman2020adversariallytrans} on the ImageNet benchmark. The horizontal axis represents the perturbation strength in training the model, e.g. "0.05" represents the robust model trained with $\ell_{2}$ perturbation $\epsilon=0.05$. The strength of perturbation has a great influence on GradNorm \cite{huang2021importance}, but little influence on other methods. Our BATS surpasses all the existing methods on different robust models.
 
\section{Related literature to OOD detection}\label{App:related}
OOD detection has received wide attention because it is critical to ensuring the reliability and safety of deep neural networks.
The literature related to OOD detection can be broadly grouped into the following themes. Our paper briefly reviews the literature related to post-hoc detection methods. Here, we provide a more comprehensive review. 

\textbf{Post-hoc Detection Methods.} Post-hoc methods focus on improving the OOD uncertainty estimation by utilizing the pre-trained classifiers rather than retraining a model, which is beneficial for adopting OOD detection in real-world scenarios and large-scale settings. In this paper, we mainly focus on the post-hoc OOD detection methods. \citet{hendrycks17baseline} observe that the maximum softmax probability of In-Distribution Examples (ID) can be higher than the Out-of-Distribution (OOD) samples and provide a simple baseline for OOD detection. ODIN \cite{ODIN} introduces a large sufficiently temperature factor and input perturbation to separate the ID and OOD examples. \citet{liu2020energy} analyze the limitations of softmax function in OOD detection and propose to use energy score as an indicator. The examples with high energy are considered as OOD examples, and vice versa. \citet{wang2021can} propose to use joint energy score which take labels into consideration to enhance the OOD detection. ReAct \cite{sun2021react} hypothesize that the OOD examples can trigger the abnormal activation of the model, and propose to clamp the activation value larger than the threshold value to improve the detection performance. \citet{Mahalanobis} use the mixture of Gaussians to model the distribution of feature representations and propose using the feature-level Mahalanobis distance instead of the model output. GradNorm \cite{huang2021importance} shows that the gradients of the categorical cross-entropy loss contains useful information for OOD detection.

\textbf{Confidence Enhancement Methods.}
To enhance the sensitivity to OOD examples, some methods propose to introduce the adversarial examples into the training process. \citet{hein2019relu} endow low confidence predictions to the examples far away from the training data through an adversarial training optimization technique. Moreover, \citet{bitterwolf2020certifiably} enforce low confidence in an l2 ball around the OOD examples. Proper data augmentation also contributes to OOD uncertainty estimation \cite{hendrycks*2020augmix,yun2019cutmix,thulasidasan2019mixup}. Some methods take advantage of a set of collected OOD examples to enhance the uncertainty estimation, which are named outlier exposure methods \cite{hendrycks2018deep,papadopoulos2021outlier,chen2021atom}. The correlations between the collected and real OOD examples can largely affect the performance of outliers exposure methods \cite{shafaei2019less}. 

\textbf{Density-based Methods.}
Directly estimating the density of the examples can be a natural approach, this kind of methods explicitly model the distribution of ID examples with probabilistic models and distinguish the OOD/ID examples through the likelihood \cite{kobyzev2020normalizing,zisselman2020deep,serra2019input,xiao2020likelihood}. However, some works show that the probabilistic models may assign higher likelihood to OOD examples than ID examples and fail to distinguish OOD/ID examples \cite{nalisnick2018deep,kirichenko2020normalizing}.

\section{Test accuracy}\label{App:testacc}

\begin{figure}[htbp]
\centering
\includegraphics[width=1.0\textwidth]{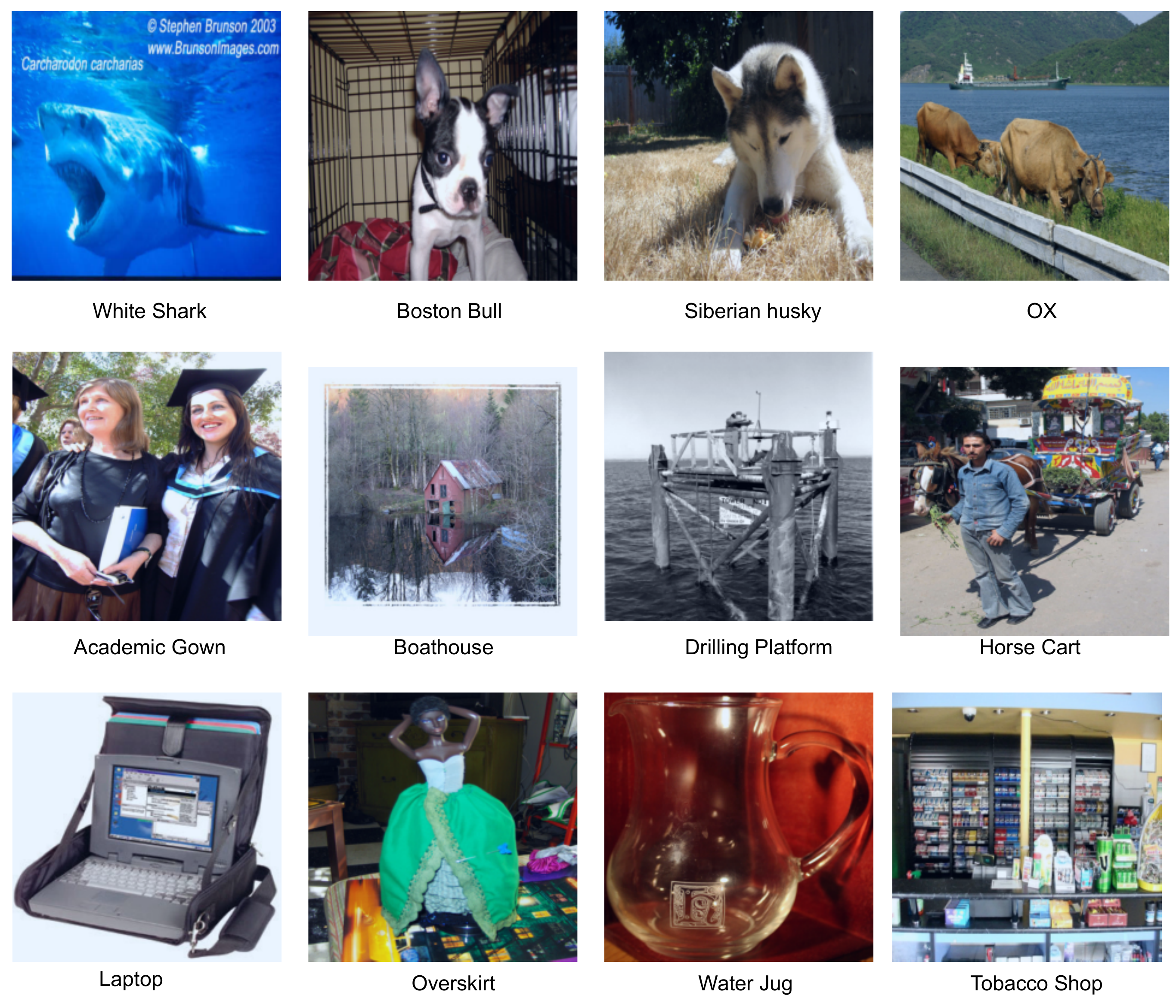}
\caption{We illustrate some examples in ImageNet that can be classified correctly by the feature rectified ResNet-50 but classified wrongly by the original ResNet-50.}
\label{img:acc_right}
\end{figure}

In this section, we show that with a proper hyper-parameter $\lambda$, our feature rectification method can slightly improve the test accuracy of the pre-trained models both on the clean images and the corrupted images. Here we set $\lambda=3$. As shown in Tab. \ref{tab:accuracy}, we evaluate the test accuracy of the normal pre-trained models and the pre-trained models with our feature rectification method on the clean images and the corrupted images. We choose some image corruption methods used in \cite{hendrycks2018benchmarking}, including salt-and-pepper noise (SP(0.2)), cropout (Crop(0.8)), JPEG compression (JPEG(50)), Gaussian blur (GB) and Gaussian noise (GN). Fig. \ref{img:acc_right} shows some examples that can be classified correctly by our feature rectified ResNet-50 but classified wrongly by the original ResNet-50.

\begin{table}[htbp]
\caption{Test accuracy on ImageNet with the pre-trained ResNet-50 and DenseNet-121. Our method rectifies the feature vector of the model and performs well on both the clean images and the corrupted images. The best results are in bold.}
\scalebox{0.91}{
\begin{tabular}{cccccccccc}
\hline \hline
Model & Method & Vanilla & SP(0.2) & Crop(0.6) & JPEG(50) & GB(2) & GB(3) & GN(0.5) & GN(1) \\ \hline
\multirow{2}{*}{RN50} & Normal & 74.548 & 40.436 & 65.488 & 58.426 & 52.762 & 49.022 & 28.638 & 10.864 \\ \cline{2-10} 
 & Ours & \textbf{74.610} & \textbf{40.506} & \textbf{65.770} & \textbf{58.454} & \textbf{52.920} & \textbf{49.312} & \textbf{28.852} & \textbf{11.014} \\ \hline \hline
\multirow{2}{*}{DN121} & Normal & 71.956 & 43.078 & 63.150 & 61.298 & 50.362 & 46.980 & 40.600 & 25.786 \\ \cline{2-10} 
 & Ours & \textbf{72.050} & \textbf{43.146} & \textbf{63.498} & \textbf{61.332} & \textbf{50.520} & \textbf{47.014} & \textbf{40.618} & \textbf{25.852} \\ \hline \hline
\end{tabular}\label{tab:accuracy}}
\end{table}

\section{BATS on other detection methods}\label{App:combine}
In our paper, we provide a concise and effective approach \textbf{BATS} to improve the performance of the existing OOD detection methods.
We mainly show the effectiveness of applying our feature rectification (BATS) on Energy Score \cite{liu2020energy}. In Tab. \ref{tab:combine} we show that out method is compatible with various OOD detection test statistics (including output-based methods \cite{hendrycks17baseline,liu2020energy,ODIN} and gradient-based method \cite{huang2021importance}) and can bring improvements to different methods. Applying our \textbf{BATS} on GradNorm can even achieve better performance than "Energy+\textbf{BATS}" but this method needs to derive the gradients of the model which cost more than "Energy+\textbf{BATS}".

\begin{figure}[htbp]  
\centering
\includegraphics[width=1.0\textwidth]{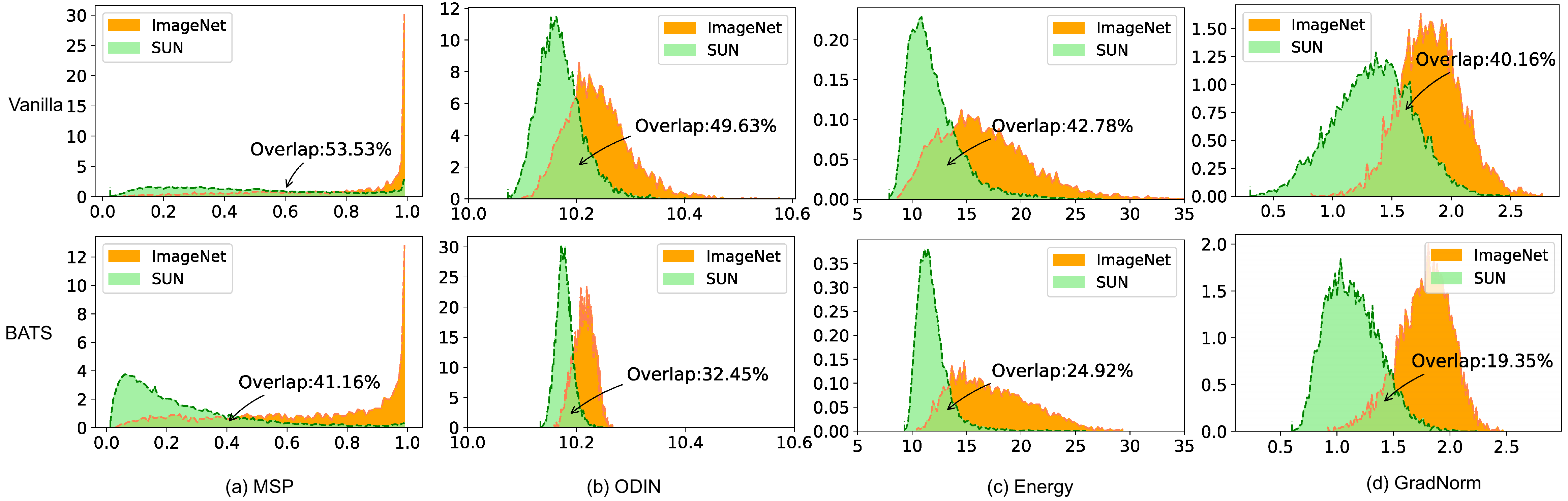}
\caption{We illustrate different OOD scores for ID (ImageNet) and OOD (SUN) examples. "Vanilla" means the original OOD detection method, and "BATS" means applying our BATS to the detection method. BATS reduces the variance of the scores and reduces the overlap between the distribution of ID and OOD examples.}
\label{img:BATS_SUN}
\end{figure}

{
OOD detection methods hope to assign higher scores to the in-distribution (ID) examples and lower scores to the out-of-distribution (OOD) examples. The advanced detection method (GradNorm \cite{huang2021importance}) can assign better scores to ID and OOD examples than the simple baseline method (MSP \cite{hendrycks17baseline}). However, there still exists an overlap in the distribution of the scores. As shown in Fig. \ref{img:BATS_SUN}, our BATS reduces the variance of the scores and makes the scores of the ID and OOD examples more separable, which can improve the performance of the OOD detection methods. We think combining our method with a better OOD score can achieve better performance. ``BATS+Energy" has already achieved state-of-the-art performance on large-scale and small-scale benchmarks.}

\begin{table}[htbp]
\centering
\caption{Applying feature rectification (\textbf{BATS}) and ReAct \cite{sun2021react} on different OOD detection methods. The best results are in bold.}
\scalebox{0.715}{
\begin{tabular}{cccccccccccc}
\hline \hline
\multirow{2}{*}{Model} & \multirow{2}{*}{Method} & \multicolumn{2}{c}{iNaturalist} & \multicolumn{2}{c}{SUN} & \multicolumn{2}{c}{Places} & \multicolumn{2}{c}{Textures} & \multicolumn{2}{c}{Average} \\ \cline{3-12} 
 &  & FPR95 & AUROC & FPR95 & AUROC & FPR95 & AUROC & FPR95 & AUROC & FPR95 & AUROC \\ \hline
\multirow{12}{*}{RN50} & MSP & 51.44 & 88.17 & 72.04 & 79.95 & 74.34 & 78.84 & 54.90 & 78.69 & 63.18 & 81.41 \\
 & MSP+ReAct & 44.90 & 91.68 & 60.86 & 86.22 & 64.95 & 84.48 & 62.06 & 85.46 & 58.19 & 86.96 \\
 & MSP+\textbf{BATS} & 35.79 & 93.56 & 56.97 & 88.08 & 63.24 & 85.35 & 55.14 & 87.93 & 52.79 & 88.73 \\ \cline{2-12} 
 & ODIN & 41.07 & 91.32 & 64.63 & 84.71 & 68.36 & 81.95 & 50.55 & 85.77 & 56.15 & 85.94 \\
 & ODIN+ReAct & 32.10 & 93.84 & 45.14 & 90.34 & 52.48 & 87.92 & 45.07 & 87.95 & 43.70 & 90.01 \\
 & ODIN+\textbf{BATS} & 25.43 & 95.44 & 40.12 & 92.28 & 50.57 & 88.87 & 36.67 & 92.42 & 38.20 & 92.25 \\ \cline{2-12} 
 & Energy & 46.65 & 91.32 & 61.96 & 84.88 & 67.97 & 82.21 & 56.06 & 84.88 & 58.16 & 85.82 \\
 & Energy+ReAct & 17.77 & 96.70 & 25.15 & 94.34 & 34.64 & 91.92 & 51.31 & 88.83 & 32.22 & 92.95 \\
 & Energy+\textbf{BATS} & 12.57 & 97.67 & 22.62 & 95.33 & 34.34 & 91.83 & 38.90 & 92.27 & 27.11 & 94.28 \\ \cline{2-12} 
 & GradNorm & 23.73 & 93.97 & 42.81 & 87.26 & 55.62 & 81.85 & 38.15 & 87.73 & 40.08 & 87.70 \\
 & GradNorm+ReAct & 12.95 & 97.74 & 26.41 & 94.85 & 38.44 & 91.70 & 29.55 & 93.78 & 26.84 & 94.52 \\
 & GradNorm+\textbf{BATS} & \textbf{10.01} & \textbf{98.23} & \textbf{18.87} & \textbf{96.42} & \textbf{32.45} & \textbf{92.78} & \textbf{24.79} & \textbf{95.28} & \textbf{21.53} & \textbf{95.68} \\ \hline \hline
\multirow{12}{*}{MNet} & MSP & 63.09 & 85.71 & 79.67 & 76.01 & 81.47 & 75.51 & 75.12 & 76.49 & 74.84 & 78.43 \\
 & MSP+ReAct & 65.42 & 86.90 & 81.09 & 76.09 & 81.68 & 75.68 & 69.93 & 81.34 & 74.53 & 80.00 \\
 & MSP+\textbf{BATS} & 49.77 & 90.60 & 70.75 & 80.66 & 74.66 & 78.45 & 57.61 & 85.61 & 63.20 & 83.83 \\ \cline{2-12} 
 & ODIN & 45.61 & 91.33 & 63.03 & 83.44 & 70.01 & 80.85 & 52.45 & 85.61 & 57.78 & 85.31 \\
 & ODIN+ReAct & 41.90 & 92.36 & 68.29 & 82.82 & 71.96 & 81.00 & 43.37 & 89.76 & 56.38 & 86.49 \\
 & ODIN+\textbf{BATS} & 29.15 & 94.66 & 58.54 & 85.38 & 65.60 & 82.24 & 35.96 & 91.42 & 47.31 & 88.43 \\ \cline{2-12} 
 & Energy & 49.52 & 91.10 & 63.06 & 84.42 & 69.24 & 81.42 & 58.16 & 84.88 & 60.00 & 85.46 \\
 & Energy+ReAct & 37.08 & 93.41 & 53.13 & 86.04 & 54.15 & 83.31 & 42.45 & 89.42 & 46.70 & 88.05 \\
 & Energy+\textbf{BATS} & 31.56 & 94.33 & 41.68 & 90.21 & 52.43 & 86.26 & 38.69 & 90.76 & 41.09 & 90.39 \\ \cline{2-12} 
 & GradNorm & 33.70 & 92.46 & 42.15 & 89.65 & 56.56 & 83.93 & 34.95 & 90.99 & 41.84 & 89.26 \\
 & GradNorm+ReAct & 25.85 & 95.03 & 38.94 & 91.42 & 52.94 & 86.74 & 18.85 & 95.75 & 34.15 & 92.24 \\
 & GradNorm+\textbf{BATS} & \textbf{21.51} & \textbf{95.89} & \textbf{30.97} & \textbf{93.19} & \textbf{46.94} & \textbf{88.08} & \textbf{17.71} & \textbf{95.97} & \textbf{29.28} & \textbf{93.28} \\ \hline \hline
\end{tabular}\label{tab:combine}}
\end{table}

\section{The feature distribution on different channels}\label{App:channels}

\begin{figure}[htbp]
\centering
\includegraphics[width=0.7\textwidth]{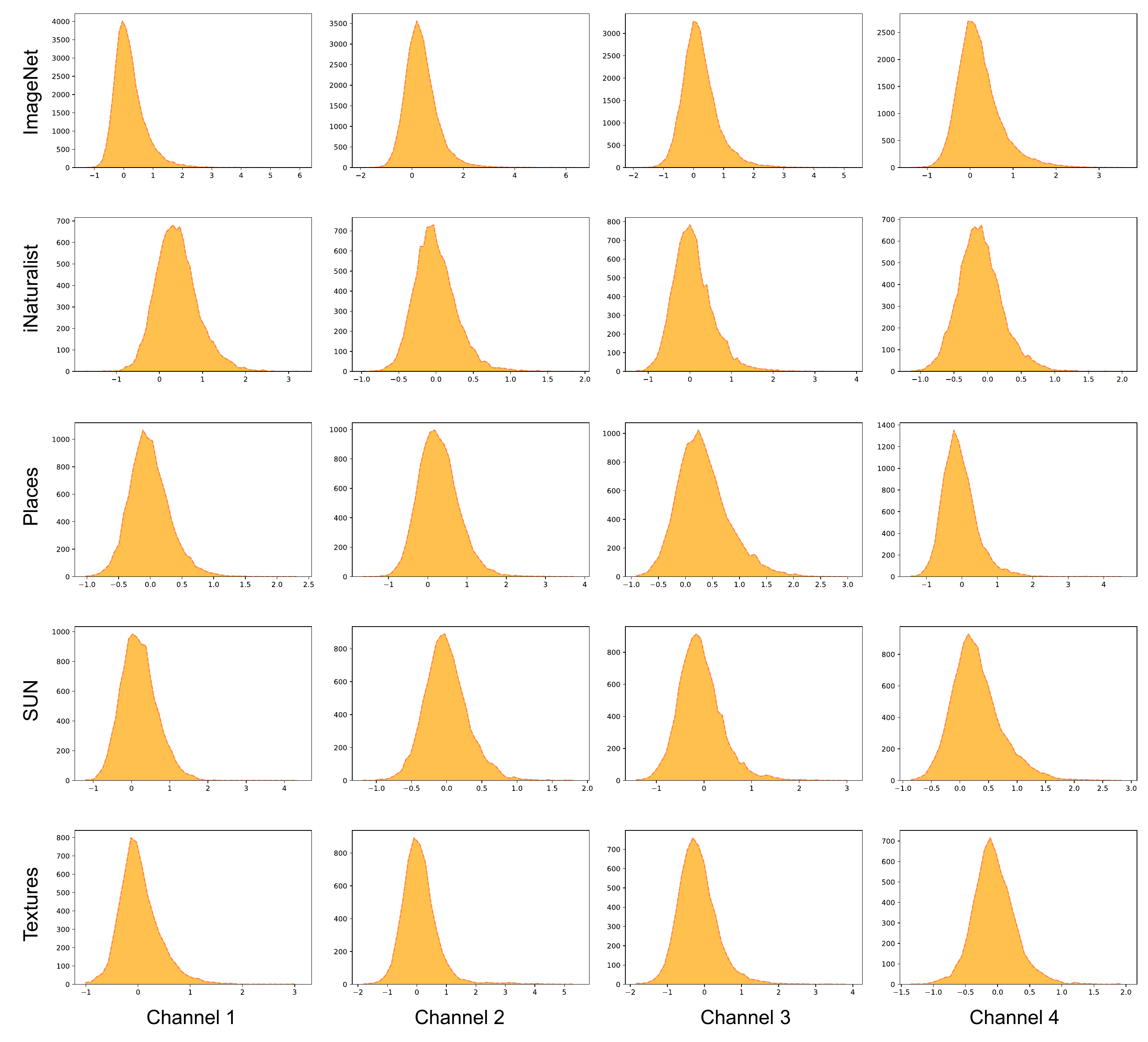}
\caption{The feature distribution on different channels of in-distribution examples (ImageNet) and the out-of-distribution examples on ResNet-50. We randomly choose four channels of the features extracted by the penultimate layer.}
\label{img:channels}
\end{figure}

Fig. \ref{img:channels} illustrates the feature distribution on different channels of in-distribution examples (ImageNet) and the out-of-distribution examples on ResNet-50. These features are extracted by the last convolution block. We name the region where features are concentrated as the typical set of features. These regions receive more attention during training, and the model is more familiar with the features in these regions than those in extreme regions.
For better visual presentation, we illustrate features before ReLU.

\section{Detection on natural adversarial examples}\label{App:NAE}
\begin{figure}[htbp]
\centering
\includegraphics[width=0.9\textwidth]{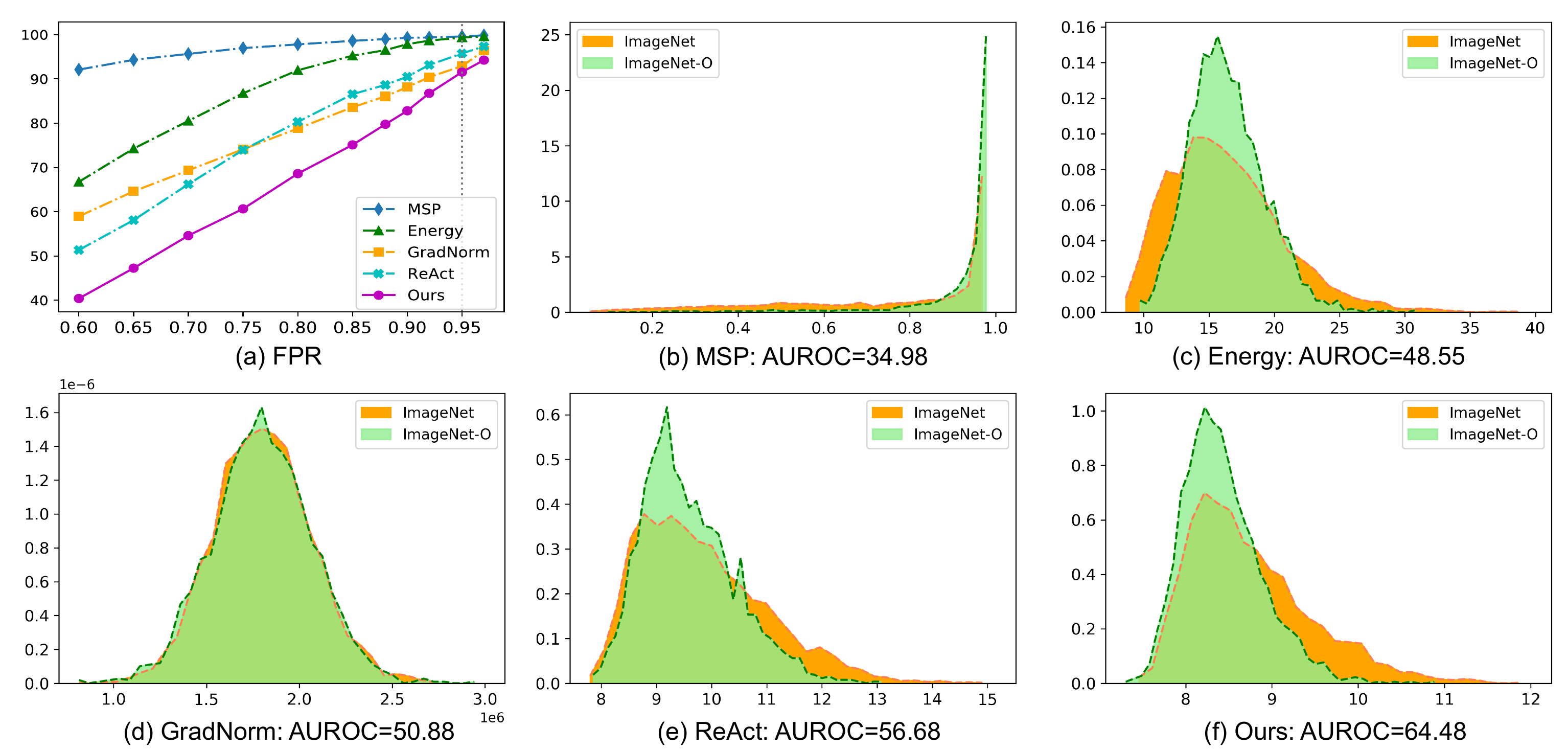}
\caption{(a) The FPR(1-$\alpha$) for different method on Imagenet (lower is better). The model is ResNet-50 and the OOD dataset is ImageNet-O. (b)-(f) The frequency histogram of the scores for ImageNet and ImageNet-O.}
\label{img:imagenet-o}
\end{figure}

In our paper, we follow the settings of the existing research, choosing iNaturalist, Places, SUN and Textures as out-of-distribution datasets and choosing ImageNet-1k as the in-distribution dataset. Here we consider a much more challenging task: detecting natural adversarial examples \cite{hendrycks2021nae}. \citet{hendrycks2021nae} introduce natural adversarial examples ImageNet-O, which are naturally occurring examples in the real world but significantly degrade the deep model performance. We use the ImageNet-O \cite{hendrycks2021nae} which contains anomalies of unforeseen classes as the out-of-distribution examples. Fig. \ref{img:imagenet-o} shows that our method surpasses the existing methods by a large margin in FPR(1-$\alpha$) with different significance levels. Our method significantly improve the AUROC from 56.68$\%$ to 64.48$\%$.

\section{Class activation mapping}\label{App:activation}

\begin{figure}[htbp]
\centering
\includegraphics[width=0.95\textwidth]{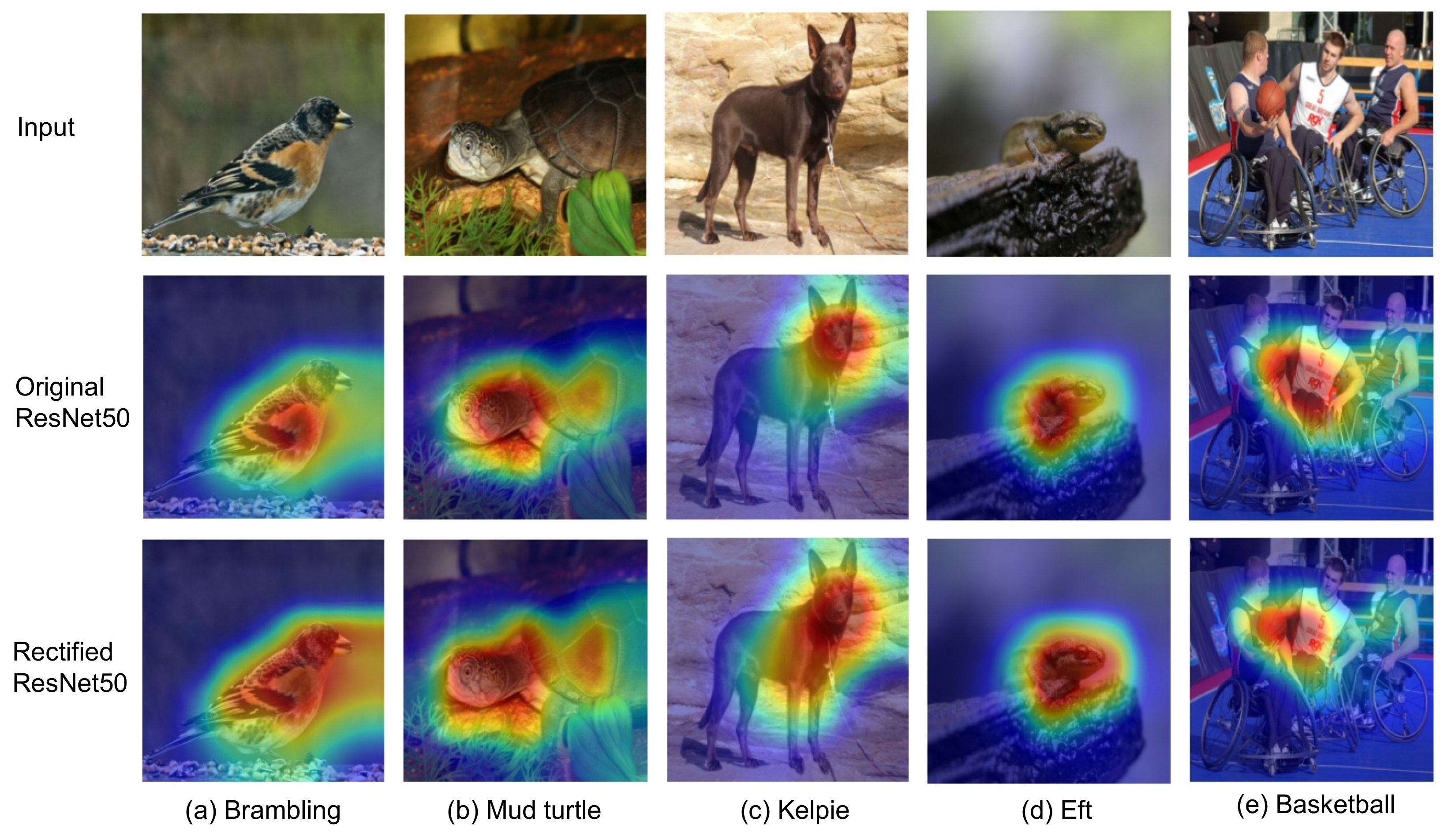}
\caption{We draw the heat maps to explain which parts of the image dominate the model decision through Smooth Grad-CAM++\cite{omeiza2019smooth}. The heat map of our rectified model for each image aligns better with the objects in the image than that of the original model. }
\label{img:cam}
\end{figure}

In this section, we use the Smooth Grad-CAM++\cite{omeiza2019smooth} to generate the heat map for different images. As shown in Fig. \ref{img:cam}, the heat map of our rectified model aligns better with the objects in the image than that of the original model. We use the pre-trained ResNet-50 in PyTorch \cite{pytorch}. The heat map of our rectified model for Fig. \ref{img:cam}(b) (the mud turtle) shows that the head of the turtle dominates the decision while the original model pays more attention to the neck of the turtle. The rectified model takes more object-relevant parts into consideration, which may contribute to its slightly better test accuracy (in Appendix \ref{App:testacc}).

\begin{figure}[htbp] 
\centering
\includegraphics[width=0.99\textwidth]{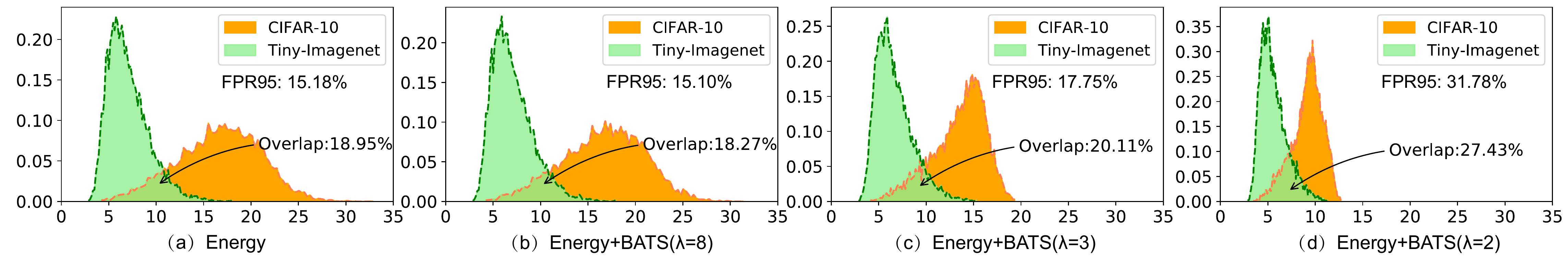}
\caption{The distribution of the scores for ID (CIFAR-10) and OOD examples (Tiny-ImageNet) on ResNet-18. Choosing a proper $\lambda$, BATS can reduce the overlap between the ID and OOD examples and reduce the FPR95, while a small $\lambda$ hinders the performance of OOD detection.}
\label{img:Tiny_Imagenet}
\end{figure}

{
\section{Additional Analysis for Performance Degradation Case in Tab.2}\label{App:Degradation}
 In Tab. \ref{tab:CIFAR}, we show that the average performance of our \textbf{BATS} surpasses the baseline method Energy, but the performance degrades in the case using Tiny-Imagenet as the OOD dataset.
 We hypothesize that this performance degradation is due to the bias introduced by BATS. By truncating the features, BATS can reduce the variance of the in-distribution examples which benefits the estimation of the reject region but inherently cause some information loss which may reduce the performance of the pre-trained models. 
 
 To validate our hypothesis, we tune the bias-variance trade-off by the hyperparameter $\lambda$.
 As shown in Fig. \ref{img:Tiny_Imagenet}, BATS can indeed reduce the variance of the OOD scores. With a proper $\lambda$, BATS can reduce the overlap between the ID and OOD examples and reduce the FPR95, while a small $\lambda$ hinders the performance of OOD detection. For example, using larger $\lambda=8$, BATS can achieve better FPR95 performance 15.10$\%$ on detecting Tiny-Imagenet using ResNet-18, which is 2.65$\%$ better than $\lambda=3$ in our Tab. \ref{tab:CIFAR}. For the practicability of our method, we set the same hyperparameter to test different OOD datasets, without adjusting for specific OOD datasets.
 }
 
\begin{figure}[htbp] 
\centering
\includegraphics[width=0.99\textwidth]{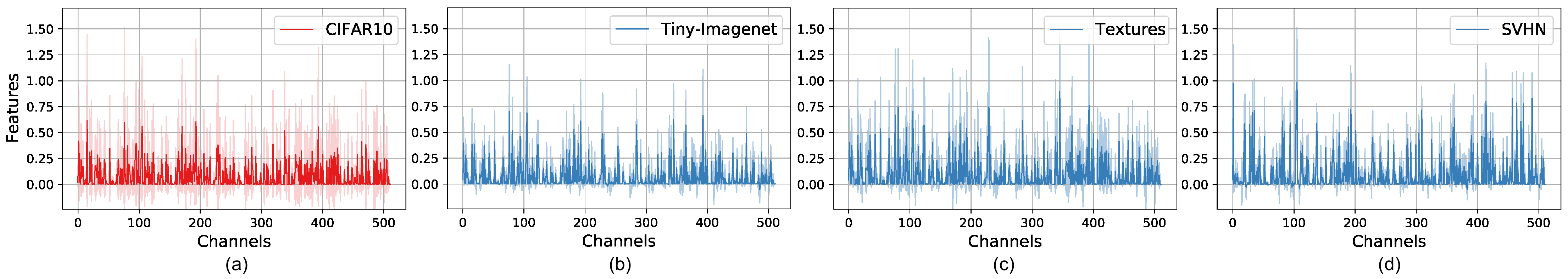}
\caption{The distribution of the features of the in-distribution dataset (a) and out-of-distribution datasets (b-d) on different channels. We use the WideResNet-28-10 to extract the features. The mean and standard deviation are shown by the solid line and shaded area, respectively. Compared to other datasets, the mean value of the features in different channels of the Tiny-Imagenet is smaller and has a smaller standard deviation.}
\label{img:cifar10_channels}
\end{figure}

 {
 \section{The difference between BATS and ReAct}
 
 The similarity between our \textbf{BATS} and ReAct is that these methods are used to improve the performance of the existing OOD scores. As follows, we discuss the difference between \textbf{BATS} and ReAct from three aspects.

 First, the motivation between our \textbf{BATS} and ReAct is different. ReAct hypothesizes that the mean activation of OOD data has significantly larger variations across units and is biased towards having sharp positive values, while the activation of the ID data is well-behaved with a near-constant mean and standard deviation. Thus, ReAct thinks that the truncation can rectify the activation of the OOD examples and preserve the activation of in-distribution data.
 However, as shown in Fig. \ref{img:cifar10_channels}, the mean activation of OOD data does not always have significantly larger variations than the ID data, which means this hypothesis does not always hold. 
 The distribution of the deep features after batch normalization is consistent with the Gaussian distribution. Our \textbf{BATS} hypothesizes that deep models may be hard to model the extreme features but can provide reliable estimations on the typical features. This is because extreme features are exposed to the training process with a low probability. We propose to rectify the features into the typical set and calculate the OOD scores with the typical features.

 Second, the mathematical analysis between our BATS and ReAct is different. ReAct theoretically analyze that if the OOD activations are more positively skewed, their operation reduces mean OOD activations more than ID activations.
 We analyze the benefit of \textbf{BATS} from the perspective of the bias-variance trade-off. BATS can reduce the variance of the deep features, which contributes to constraining the uncertainty of the test static $T(x; f)$ and improving the estimation accuracy of the reject region. Our method hopes to estimate the reject region better, and we do not assume whether OOD data is positively skewed.

 Third, our method surpasses the ReAct in both the large-scale benchmark (ImageNet) and the small-scale benchmark (CIFAR).
 
 }

{
\section{Selecting features' typical set without assistance of BN}
In our paper, we mainly analyze that rectifying the features in the typical set can improve the performance of the existing OOD scores. We provide a concise and effective method to select the typical set with the assistance of the BN layers and achieve a state-of-the-art performance among post-hoc methods on a suite of OOD detection benchmarks. 

Here, we provide another method to select the features' typical set, which directly uses a set of training images to estimate the mean $\mu$ and the standard deviation $\sigma$ of the features (extracted by the penultimate layer of the model) at each dimension. Then we rectify the features into the interval $[\mu-\lambda*\sigma, \mu+\lambda*\sigma]$ and use these typical features to calculate the OOD scores. We named this method as \textbf{T}ypical \textbf{F}eature \textbf{E}stimated \textbf{M}ethod (TFEM). This method does not require the BN layers in the model but needs to use a set of training images. 

In Tab. \ref{othermethod}, we compare the OOD detection performance of the OOD detection methods with and without our TFEM. In this experiment, we randomly choose 1500 images from the training dataset of the ImageNet. The $\lambda$ is set to 1. The experiment is performed on the ImageNet benchmark. The models are pre-trained ResNet-50 and ViT. ViT (Vision Transformer) \cite{dosovitskiy2020imageVIT} is a transformer-based image classification model which treats images as sequences of patches and does not have BN layers. We use the officially released ViT-B/16 model, which is pre-trained on ImageNet-21K and fine-tuned on ImageNet-1K. Rectifying the features into the typical set with TFEM can greatly improve the performance of the existing OOD detection methods both on the model with BN layers (ResNet-50) and the model without BN layers (ViT).

This experiment demonstrates the effectiveness of the typical features in OOD detection, which is consistent with the analysis in our paper.
We believe there exists a method that can estimate the features' typical set better. In this paper, BATS has already established state-of-the-art performance on both the large-scale and small-scale OOD detection benchmarks.
}

\begin{table}[htbp]  
\caption{Using TFEM to select features' typical set. We use the pre-trained ResNet-50 and ViT-B/16 to detect the OOD examples. The best results are in bold.}
\scalebox{0.7}{
\begin{tabular}{cccccccccccc}
\hline
\multirow{2}{*}{Model} & \multirow{2}{*}{Method} & \multicolumn{2}{c}{iNaturalist} & \multicolumn{2}{c}{SUN} & \multicolumn{2}{c}{Places} & \multicolumn{2}{c}{Textures} & \multicolumn{2}{c}{Average} \\
 &  & FPR95 & AUROC & FPR95 & AUROC & FPR95 & AUROC & FPR95 & AUROC & FPR95 & AUROC \\ \hline
\multirow{8}{*}{ViT} & MSP & 16.15 & 96.37 & 56.56 & 85.18 & 59.39 & 84.62 & 50.99 & 84.68 & 45.77 & 87.71 \\
 & MSP+\textbf{TFEM} & \textbf{4.10} & \textbf{99.09} & \textbf{40.62} & \textbf{90.97} & \textbf{47.43} & \textbf{89.20} & \textbf{39.70} & \textbf{89.06} & \textbf{32.96} & \textbf{92.08} \\ \cline{2-12} 
 & ODIN & 13.90 & 96.88 & 43.91 & 88.89 & 52.19 & 85.90 & 42.36 & 88.35 & 38.09 & 90.01 \\
 & ODIN+\textbf{TFEM} & \textbf{6.45} & \textbf{98.78} & \textbf{35.44} & \textbf{92.39} & \textbf{45.36} & \textbf{89.34} & \textbf{43.07} & \textbf{88.61} & \textbf{32.58} & \textbf{92.28} \\ \cline{2-12} 
 & Energy & 5.26 & 98.62 & 40.81 & 90.80 & 48.75 & 88.44 & 34.06 & 91.25 & 32.22 & 92.28 \\
 & Energy+\textbf{TFEM} & \textbf{1.48} & \textbf{99.68} & \textbf{29.19} & \textbf{93.84} & \textbf{40.12} & \textbf{91.22} & \textbf{30.44} & \textbf{92.17} & \textbf{25.31} & \textbf{94.23} \\ \cline{2-12} 
 & GradNorm & 5.14 & 98.35 & 42.06 & 89.26 & 49.21 & 86.63 & 35.57 & 89.27 & 33.00 & 90.88 \\
 & GradNorm+\textbf{TFEM} & \textbf{1.50} & \textbf{99.66} & \textbf{28.86} & \textbf{93.88} & \textbf{40.04} & \textbf{91.27} & \textbf{30.69} & \textbf{92.08} & \textbf{25.27} & \textbf{94.22} \\ \hline
 \multirow{8}{*}{ResNet50} & MSP & 51.44 & 88.17 & 72.04 & 79.95 & 74.34 & 78.84 & \textbf{54.90} & 78.69 & 63.18 & 81.41 \\
 & MSP+\textbf{TFEM} & \textbf{38.50} & \textbf{92.77} & \textbf{66.53} & \textbf{84.47} & \textbf{70.59} & \textbf{82.13} & 58.40 & \textbf{86.71} & \textbf{58.51} & \textbf{86.52} \\ \cline{2-12} 
 & ODIN & 41.07 & 91.32 & 64.63 & 84.71 & 68.36 & 81.95 & 50.55 & 85.77 & 56.15 & 85.94 \\
 & ODIN+\textbf{TFEM} & \textbf{28.40} & \textbf{94.67} & \textbf{52.34} & \textbf{89.47} & \textbf{62.13} & \textbf{85.14} & \textbf{37.27} & \textbf{92.35} & \textbf{45.04} & \textbf{90.41} \\ \cline{2-12} 
 & Energy & 46.65 & 91.32 & 61.96 & 84.88 & 67.97 & 82.21 & 56.06 & 84.88 & 58.16 & 85.82 \\
 & Energy+\textbf{TFEM} & \textbf{20.29} & \textbf{96.24} & \textbf{53.98} & \textbf{86.85} & \textbf{43.37} & \textbf{90.90} & \textbf{38.24} & \textbf{92.22} & \textbf{38.97} & \textbf{91.55} \\ \cline{2-12} 
 & GradNorm & 23.73 & 93.97 & 42.81 & 87.26 & 55.62 & 81.85 & 38.15 & 87.73 & 40.08 & 87.70 \\
 & GradNorm+\textbf{TFEM} & \textbf{11.88} & \textbf{97.83} & \textbf{26.24} & \textbf{95.00} & \textbf{40.46} & \textbf{90.77} & \textbf{25.05} & \textbf{94.85} & \textbf{25.91} & \textbf{94.61} \\ \hline
\end{tabular}\label{othermethod}
}
\end{table}

 {
 \section{Comparison between BATS and two latest detection methods}
 In this section, we compare our \textbf{BATS} with two latest OOD detection methods KNN \cite{sun2022knn} and ViM \cite{wang2022vim}. KNN is a nearest-neighbor-based OOD detection method, which computes the k-th nearest neighbor (KNN) distance between the embedding of test input and the embeddings of the training set to determine if the input is OOD or not. ViM combines the class-agnostic score from feature space and the In-Distribution class-dependent logits to calculate the OOD score. As shown in Tab. \ref{concurrent_exp}, our \textbf{BATS} outperforms the existing methods by a large margin. KNN explores and demonstrates the efficacy of the non-parametric nearest-neighbor distance for OOD detection, but its performance is worse than GradNorm and ReAct. ViM performs well on the OOD dataset Textures, but when using SUN as the OOD dataset, its performance is even worse than the simple baseline MSP.
}
\begin{table}[htbp] 
\caption{OOD detection performance comparison on ResNet-50 on the ImageNet benchmark. All methods are post hoc and can be directly used for pre-trained models. The best results are in Bold.} 
\scalebox{0.8}{
\begin{tabular}{ccccccccccc}
\hline
\multirow{2}{*}{Method} & \multicolumn{2}{c}{iNaturalist} & \multicolumn{2}{c}{SUN} & \multicolumn{2}{c}{Places} & \multicolumn{2}{c}{Textures} & \multicolumn{2}{c}{Average} \\ \cline{2-11} 
 & FPR95 & AUROC & FPR95 & AUROC & FPR95 & AUROC & FPR95 & AUROC & FPR95 & AUROC \\ \hline
MSP \cite{hendrycks17baseline} & 51.44 & 88.17 & 72.04 & 79.95 & 74.34 & 78.84 & 54.90 & 78.69 & 63.18 & 81.41 \\
ODIN \cite{ODIN}& 41.07 & 91.32 & 64.63 & 84.71 & 68.36 & 81.95 & 50.55 & 85.77 & 56.15 & 85.94 \\
Energy \cite{liu2020energy}& 46.65 & 91.32 & 61.96 & 84.88 & 67.97 & 82.21 & 56.06 & 84.88 & 58.16 & 85.82 \\ 
GradNorm \cite{huang2021importance}& 23.73 & 93.97 & 42.81 & 87.26 & 55.62 & 81.85 & 38.15 & 87.73 & 40.08 & 87.70 \\
ReAct \cite{sun2021react}& 17.77 & 96.70 & 25.15 & 94.34 & 34.64 & 91.92 & 51.31 & 88.83 & 32.22 & 92.95 \\
KNN \cite{sun2022knn}& 59.00 & 86.47 & 68.82 & 80.72 & 76.28 & 75.76 & \textbf{11.77} & \textbf{97.07} & 53.97 & 85.01 \\
ViM \cite{wang2022vim}& 77.34 & 86.46 & 90.71 & 73.80 & 89.64 & 72.15 & 16.63 & 96.37 & 68.58 & 82.20 \\
BATS(Ours) & \textbf{12.57} & \textbf{97.67} & \textbf{22.62} & \textbf{95.33} & \textbf{34.34} & \textbf{91.83} & 38.90 & 92.27 & \textbf{27.11} & \textbf{94.28} \\ \hline
\end{tabular}\label{concurrent_exp}}
\end{table}

\begin{figure}[htbp] 
 \vskip 0.2in
 \centering
 \centerline{\includegraphics[width=0.85\linewidth]{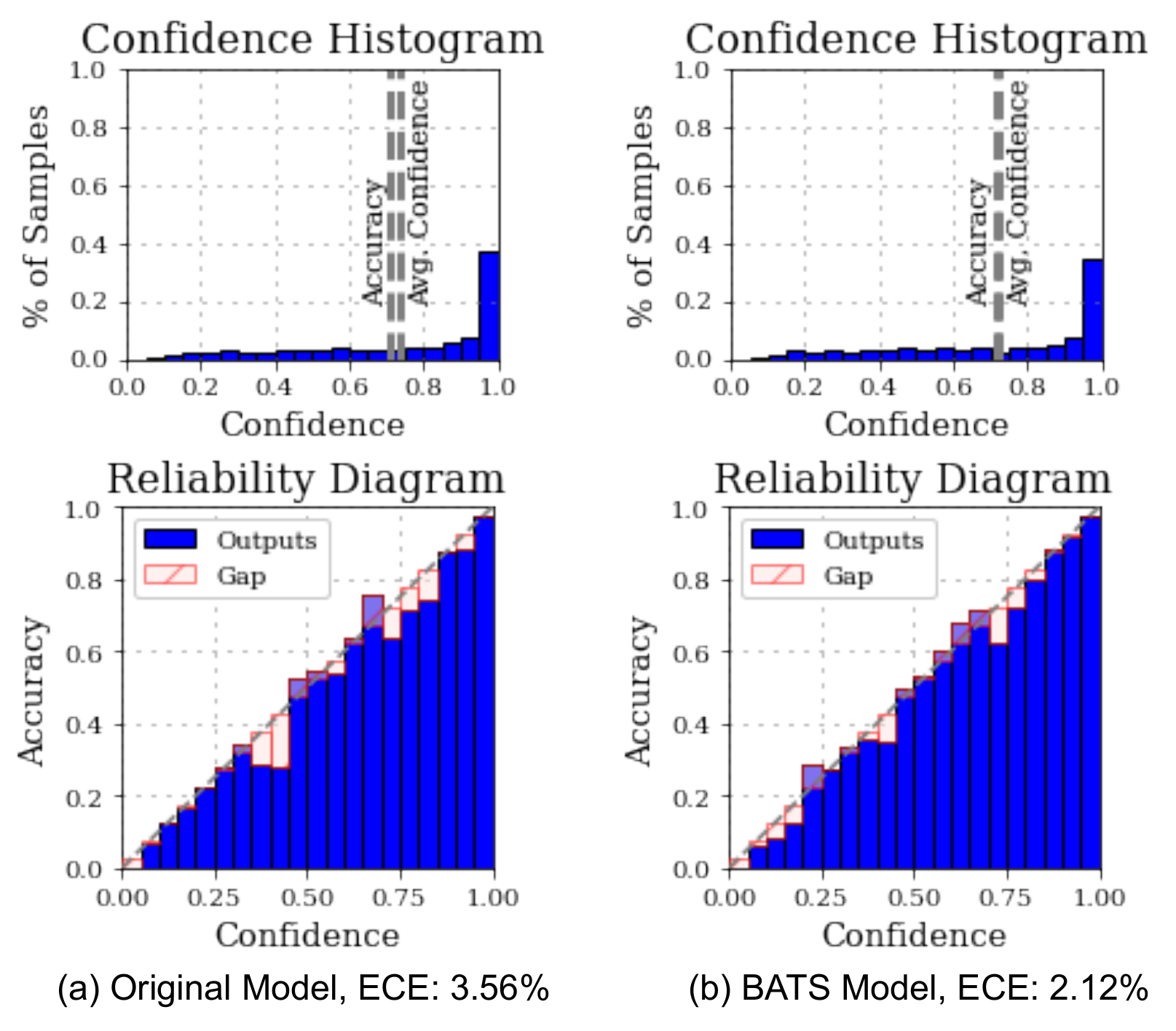}}
 \caption{We draw the reliability diagram and the confidence histogram of the pre-trained ResNet-50 (a) and the ResNet-50 with our BATS (b) on ImageNet.}
 \label{fig:calibration}
 \vskip -0.2in
 \end{figure}

{
\section{Benefits of BATS on calibration}

The outputs of a classifier are often interpreted as the predictive confidence that this class was identified. Deep neural networks are often not calibrated which means that the confidence always does not align with the misclassification rate. Expected Calibration Error (ECE) is a metric to measure the calibration of a classifier. For a perfectly calibrated classifier, the ECE value will be zero. 

We use the reliability diagram to find out how well the classifier is calibrated in Fig. ~\ref{fig:calibration}. The model's predictions are divided into bins based on the confidence value of the target class, here, we choose 20 bins. The confidence histogram shows how many test examples are in each bin. Two vertical lines represent the accuracy and average confidence, and the closer these two lines are, the better the model calibration is.  
BATS can improve the calibration of the pre-trained model and reduce the ECE of the pre-trained model from 3.56$\%$ to 2.12$\%$.

}

\end{document}

%% file: xcl_math.tex
\usepackage{amsmath,amssymb}
\usepackage{graphicx}
\usepackage{color}
\usepackage{bm}
\usepackage{algorithm}
\usepackage{algorithmic}
\usepackage{wrapfig}
\usepackage{hyperref}
\usepackage{url}

\newcommand{\benr}{\begin{eqnarray}}
\newcommand{\eenr}{\end{eqnarray}}
\newcommand{\benrr}{\begin{eqnarray*}}
\newcommand{\eenrr}{\end{eqnarray*}}
\newcommand{\ben}{\begin{equation}}
\newcommand{\een}{\end{equation}}
\newcommand{\benn}{\begin{equation*}}
\newcommand{\eenn}{\end{equation*}}

\newcommand{\cD}{\mathcal D}

\newcommand{\cH}{\mathcal H}

\newcommand{\cR}{\mathcal R}

\newcommand{\cX}{\mathcal X}
\newcommand{\cY}{\mathcal Y}

\def\rvx{{\mathbf{x}}}

\def\rvz{{\mathbf{z}}}

\newcommand{\E}{\mathbb{E}}

\newcommand{\R}{\mathbb R}

\newcommand{\rarrow}{\rightarrow}

